\documentclass[a4paper,fleqn]{cas-dc}

\usepackage[numbers]{natbib}
\usepackage{subfigure}
\usepackage{float}
\usepackage{stfloats}
\usepackage{balance}

\def\tsc#1{\csdef{#1}{\textsc{\lowercase{#1}}\xspace}}
\tsc{WGM}
\tsc{QE}
\tsc{EP}
\tsc{PMS}
\tsc{BEC}
\tsc{DE}

\begin{document}
\let\WriteBookmarks\relax
\def\floatpagepagefraction{1}
\def\textpagefraction{.001}

\shorttitle{}

\shortauthors{F.Shuang et~al.}

\title [mode = title]{OppLoD: the Opponency based Looming Detector, Model Extension of Looming Sensitivity from LGMD to LPLC2}                      

\tnotetext[1]{This research was funded in part by the Bagui Scholar Program of Guangxi; in part by the National Natural Science Foundation of China, grant numbers 61720106009, 61773359, and 62206065; and in part by the European Union’s Horizon 2020 research and innovation program under the Marie Sklodowska-Curie grant agreement No 778062 ULTRACEPT.}


%
\author[1]{Feng Shuang}





\credit{Conceptualization of this study, Methodology, Software}

\address[1]{Guangxi Key Laboratory of Intelligent Control and Maintenance of Power Equipment (LIPE), School of Electrical Engineering, Guangxi University, Nanning City, Guangxi PR., China.}

\author[1]{Yanpeng Zhu}[style=chinese]
\credit{Data curation, Writing - Original draft preparation}

\author[1]{Yupeng Xie}
\credit{}

\author[1]{Lei Zhao}

\author[1]{Quansheng Xie}

\author[1,2]{Jiannan Zhao}[orcid=0000-0003-0052-3365]
\cormark[1]


\author[2]{Shigang Yue}

\address[2]{School of Computer Science, University of Lincoln, Isac Newton Building, Lincolnshire, UK}

\cortext[cor1]{Corresponding author, Email: jzhao@gxu.edu.cn}



\begin{abstract}
Looming detection plays an important role in insect collision prevention systems. As a vital capability evolutionary survival, it has been extensively studied in neuroscience and is attracting increasing research interest in robotics due to its close relationship with collision detection and navigation. Visual cues such as angular size, angular velocity, and expansion have been widely studied for looming detection by means of optic flow or elementary neural computing research. However, a critical visual motion cue has been long neglected because it is so easy to be confused with expansion, that is radial-opponent-motion (ROM).  Recent research on the discovery of LPLC2, a ROM-sensitive neuron in Drosophila, has revealed its ultra-selectivity because it only responds to stimuli with focal, outward movement. This characteristic of ROM-sensitivity is consistent with the demand for collision detection because it is strongly associated with danger looming that is moving towards the center of the observer. Thus, we hope to extend the well-studied neural model of the lobula giant movement detector (LGMD) with ROM-sensibility in order to enhance robustness and accuracy at the same time.
In this paper, we investigate the potential to extend an image velocity-based looming detector, the lobula giant movement detector (LGMD), with ROM-sensibility. To achieve this, we propose the mathematical definition of ROM and its main property, the radial motion opponency (RMO). Then, a synaptic neuropile that analogizes the synaptic processing of LPLC2 is proposed in the form of lateral inhibition and attention. Thus, our proposed model is the first to perform both image velocity selectivity and ROM sensitivity. Systematic experiments are conducted to exhibit the huge potential of the proposed bio-inspired looming detector.
\end{abstract}




\maketitle

\section{Introduction}
\begin{figure}[t]
  \centering
  \subfigure[]{
  \includegraphics[scale=0.258]{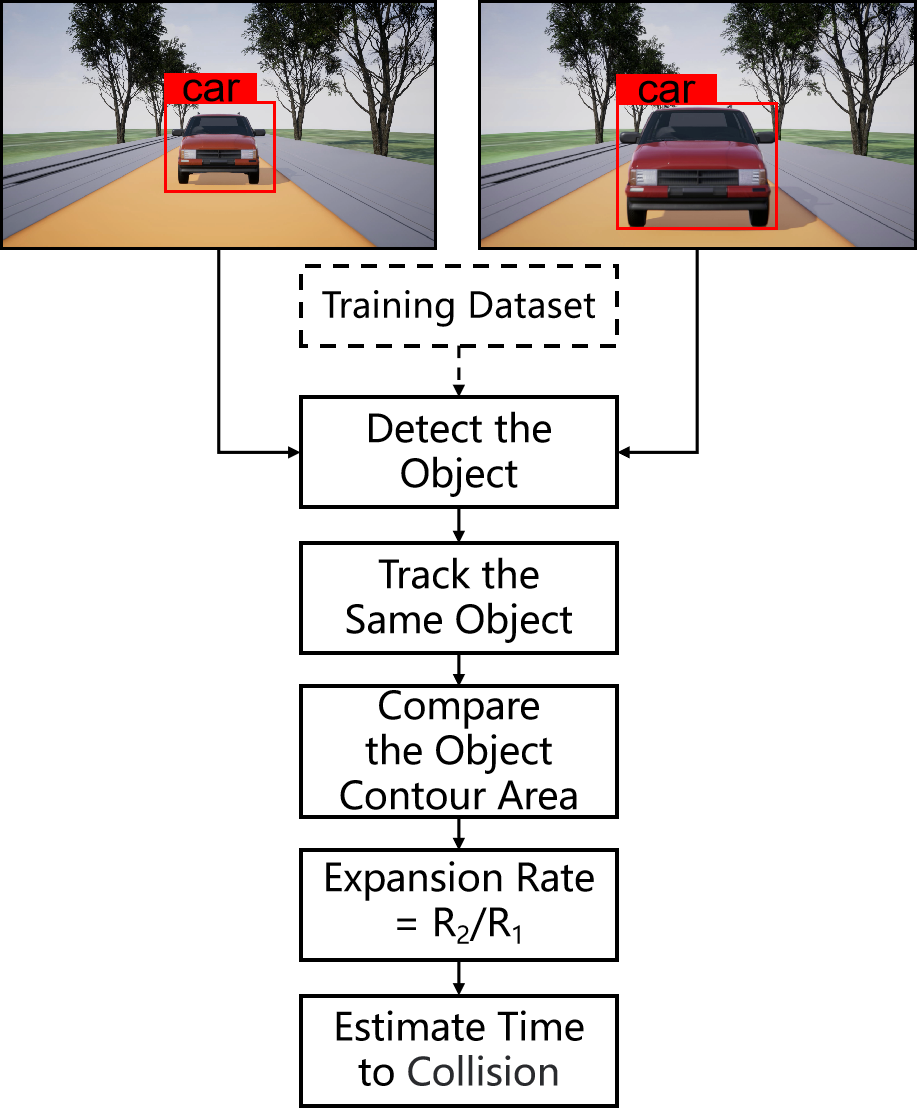}}
  \subfigure[]{
  \includegraphics[scale=0.258]{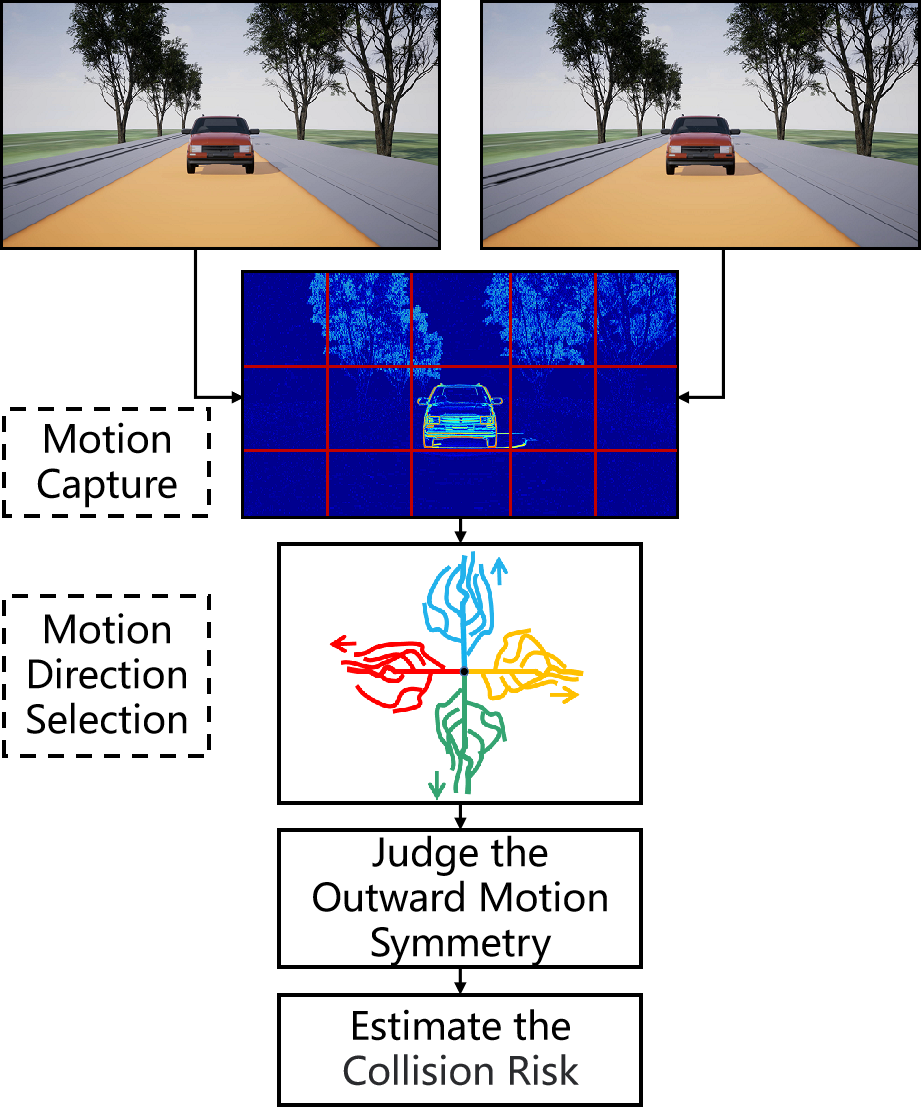}}
  \caption{Comparison of expansion detection~\cite{garcia2019Yolo} and radial motion opponency detection (RMOD). (a) The process of expansion detection. (b) The process of radial RMOD.}
  \label{fig:Comparison of expansion and ROM}
\end{figure}

Modeling the looming sensitive neuron is a thriving research topic in neural computing and bio-inspired robotics~\cite{feldstein2019impending,fu2019towards}. It promotes scientists' awareness of the animals' intelligence to predict the collision risk through the visual stimuli of looming~\cite{yan2011visual,dewell2019active,rind2016two,zhu2016fine}, which refers to the visual process of an imminent collision. Over the last decades, various visual cues including image size $\theta$, expansion rate $E_r$, image velocity $\Dot{\theta}$, etc~\cite{lee2009general,rind1996neural}, have been reported to play an important role in animals' looming sensitive neurons. Thus, computational models and realistic algorithms have been proposed to detect potential collisions via the above visual cues~\cite{zhao2021dlgmd,garcia2019Yolo,yue2006collision,1998FlowDivergence}.
However, most literature neglected the critical role of radial-opponent-motion (ROM) because it is so common to confuse it with expansion. In this paper, we propose the radial motion opponency detection (RMOD) task to select salient ROM. As shown in Fig.~\ref{fig:Comparison of expansion and ROM}, the significant difference between expansion detection and RMOD is the former requires whole object recognition, and the latter is calculated pixel-wise but only focused on outward symmetric image motion.

In order to enable artificial vehicles the ability of collision avoidance, looming detection aims to monitor the process of obstacle approaching and predict the risk of collision. Different from the traditional robotic methods in collision-free navigation which are depth and mapping-based such as Simultaneous Localization And Mapping (SLAM)\cite{jia2019survey},\cite{cadena2016past}, Artificial Potential Field (APF), heuristic search\cite{9309347},\cite{richter2018bayesian} etc., looming detection does not provide the agent the knowledge of "where it is" but the information of "where to go".
For the natural focus on visual motion, knowing "where to go" is much more computationally efficient and thus appropriate for dynamic scenarios in UAV or auto-car applications. For example, Davide\cite{falanga2020dynamic} realized differentiating the fast-moving foreground obstacles from backgrounds with an event camera and demonstrated dynamic obstacle avoidance within milliseconds. Juan\cite{rodriguez2022free} applied a similar event-based dynamic obstacle avoidance 
solution to flapping wing robots. Inspired by the locust’s looming sensitive neuron, the Lobula Giant Movement Detector(LGMD), Fu improved the LGMD and demonstrated good performance in the car crash dataset~\cite{fu2020improved}. Zhao\cite{zhao2021enhancing} proposed the D-LGMD model sensitive to image velocity $\Dot{\theta}$ and experimentally verified the performance of D-LGMD in the UAV's agile flights.
But currently, these visual motion-based methods are still perplexing to aggressive ego motions as that will cause a mass of high-speed background interference. 
Besides the above, some researchers tried to combine action strategy with planning method via learning-based neural networks. For example, DroNet~\cite{loquercio2018dronet} taught the Drone to navigate like a human rider and achieved collision-free navigation in the urban environment. This is quite an important attempt as the neural network mapped image input to avoidance steering directly. However, current deep learning usually leads to over-fitted adaption to trained environment structure and is somehow inexplicable to failures in unfamiliar scenarios. In the summary, auto-navigation in an unknown dynamic environment is still challenging.

Differently, animals can adapt to a wide range of environments because they master the secret of predicting collision from the looming process. For example, pigeons avoid prey or collide with the environment by looming sensitive neurons $\tau$ and $\rho$ ~\cite{yan2011visual,sun1998computation}, which are responsive to time-to-collision. Locusts avoid collision in a dense swarm by their looming sensitive neuron LGMD, which is responsive to image size $\theta$ and image velocity $\Dot{\theta}$. Drosophila is also reported with looming sensibility but until Nathan C. Klapoetke et.al revealed the ultra-selectivity of LPLC2 from radial motion opponency (RMO)~\cite{klapoetke2017ultra}, the visual cue of ROM just started to attract research interests~\cite{hua2022shaping,zhou2022shallow}. Different from other visual cues, radial-opponent-motion (ROM) always occurs in the visual process of observing an approaching object but may be weakened when the motion is not towards the center of the observer (as shown in Fig.2 (a)). In other words, the image motion will be more central-symmetric when an obstacle is approaching the observer's collision risky space (CRS). This indicates that opponent motion is outstandingly crucial for looming detection and this is also the reason it is called ultra-selectivity by Nathan C~\cite{klapoetke2017ultra}.

Inspired by these SOTA research~\cite {hua2022shaping,zhou2022shallow,klapoetke2017ultra}, we introduce this ultra-selectivity into the D-LGMD model~\cite{zhao2021dlgmd} with the expectation of improving robustness and accuracy. In this paper, we propose the \textbf{Opp}onency based \textbf{Lo}oming \textbf{D}etector (OppLoD), which is a synaptic computing model that combined the feature of LGMD and LPLC2 neurons, the fused model is sensitive to both radial motion opponency (RMO) and image-angular-velocity ($\Dot{\theta}$). Thus, presented a more robust preference for looming stimuli in dynamic vision. 
In summary, the contribution of this research is threefold:
\begin{itemize}
    \item Propose the definition of ROM from the aspect of dynamic vision.
    \item A lateral inhibition-based pixel-wise ROM-sensitive structure is proposed to mimic the role of the 4 layers of synaptic fans in LPLC2.
    \item The ROM-sensitive neuropile is fused with the Locust-inspired looming sensitive neural model D-LGMD so that the  OppLoD exhibits the advantages of both neurons (LGMD and LPLC2).
    \item Analytically revealed the characteristics of the proposed OppLoD with simulation and real-world video image sequences, including a UAV dataset.
\end{itemize}

\section{Related work}
Looming detection plays a vital role in collision avoidance systems for robotics and drones~\cite{yue2006collision,Fotowat2011Collision,mori2013first,salt2017obstacle}. In this section, we review the related work about looming detection from the aspect of robotic applications. Then the looming sensitive neuron of locusts (LGMD) and the heuristic neuron of Drosophila (LPLC2) are introduced respectively.

\subsection{Collision Avoidance and Looming Detection}
Looming detection is a bio-inspired concept regarding detecting potential collision risks by observing the visual process of approaching~\cite{rind1996neural,gabbiani1999computation,gabbiani1999many}. Animals are reported to exhibit the capability of predictively sensing collisions via different looming sensitive neurons~\cite{blanchard2000collision,Sergi2004collision,fu2017collision}. For example, pigeons are reported to respond to time-to-collision (TTC)~\cite{wang1992time} via looming sensitive neurons $\tau,\eta,\rho$~\cite{sun1998computation}. Each of them responds to specific visual information respectively. Locusts are reported to sense looming threats via the LGMD neuron which are sensitive to image velocity ($\Dot{\theta}$) and size ($\theta$)~\cite{rind1999seeing, Rind1984A,gabbiani1999many}. The bullfrog sense collision risks mainly depend on the size ($\theta$) of the object~\cite{nakagawa2010collision}.

Notably, Gray et.al point out that various visual sources of TTC may be occupied with different weights. The weighting of different sources has the effect of favoring the more unequivocal and reliable information and integrated to form a better sensibility to looming~\cite{gray1999monocular}. In a word, different species select different visual cues according to the different demands for survival in their natural habitat. Some of these visual cues have been applied to bio-inspired robotics for collision avoidance~\cite{fu2019review,vcivzek2017neural,hu2016bio}. But robots are required to explore the manmade environment which is more complex than a specialized habitat of one specie. Therefore, we hypothesize that fusing new principles will make the artificial looming detecting models more adaptive to complex man-made environments.

\subsection{The Locust's Looming Sensitive Neuron LGMD}
LGMD is a collision neuron found in the locust’s vision system which responds strongly to looming objects. For decades, a large number of LGMD-based neural networks have evolved. The first functional LGMD model is proposed by Rind~\cite{rind1996neural} in 1996, which selects the approaching objects by computing the expanding edges and lateral inhibition. Based on the research of Rind et al., a series of models have been developed. In order to detect colliding objects in complex background, Yue~\cite{yue2006collision} further develop the LGMD structure by adding an extra grouping layer, which enhances the clustered output. Meng~\cite{meng2010modified}introduced an improved model providing additional depth direction information which has been simplified and implemented on a Field Programmable Gate Array (FPGA). Fu~\cite{fu2019TCYB}introduced on and off pathways with splitting signals into parallel ON/OFF channels to achieve the dark object selectivity. Badia~\cite{Badia2007Blimp} use directional motion information computed by elementary-motion-detector (EMD)~\cite{reichardt1961autocorrelation} to detect expansion, but extra preprocessing is needed for their model.

Recently, Zhao~\cite{zhao2021dlgmd} proposed the D-LGMD model, which implements a spatial-temporal filter on image angular velocity, and the model is used for agile UAV flights successfully. Based on D-LGMD, a fused model with small target sensibility achieved looming powerline detection in UAV scenario~\cite{wu2022Powerline}.

\subsection{Modelling ROM-Sensitive Neuron LPLC2}
\begin{figure}[t]
  \centering
  \includegraphics[scale=1.7]{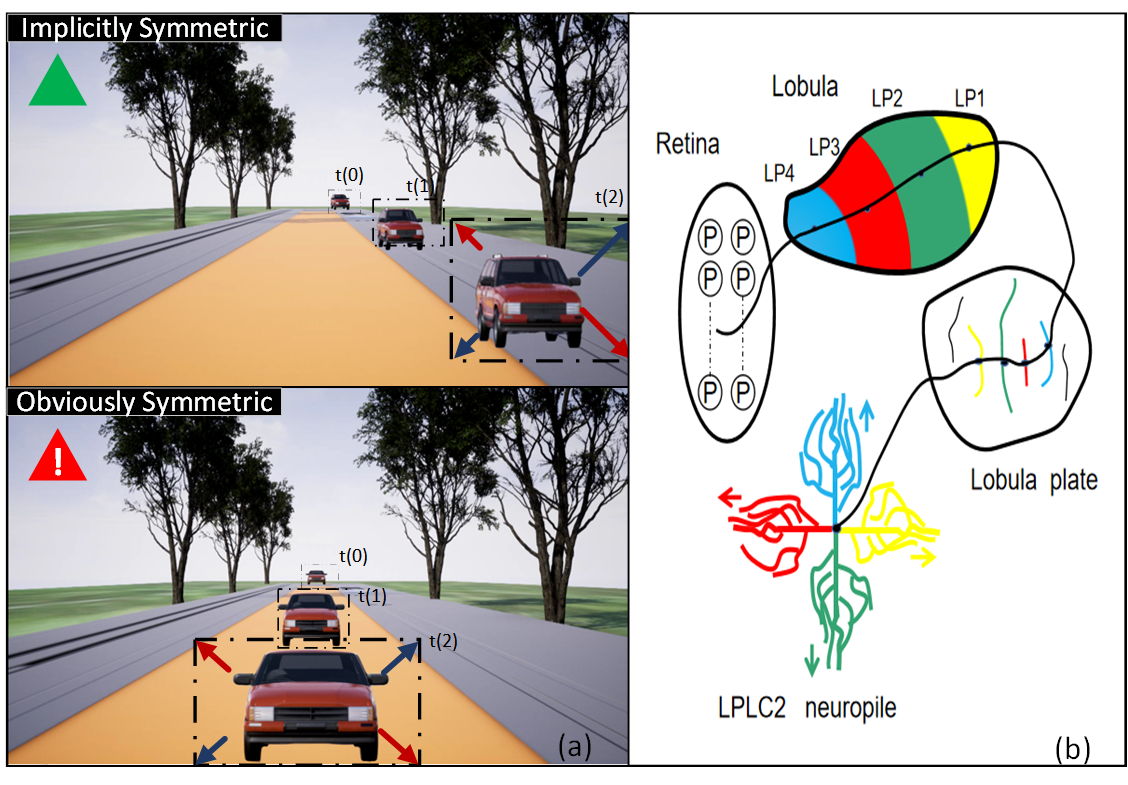}
  \caption{(a) Looming towards the center and outside the collision risky space (CRS) presents different central symmetry, which makes ROM outstandingly significant for looming detection and thus called ultra-selectivity~\cite{zhou2022shallow}. (b) The morphology of the ROM-sensitive neuron LPLC2 is the inspiration for this paper.}
  \label{fig:lplc2_model}
\end{figure}

In the visual system of the Drosophila, recent research identified an ultra-selective looming detecting neuron, namely lobula plate/lobula columnar, type II(LPLC2)~\cite{ming2016Visual,klapoetke2017ultra}, which is verified to respond strongly to outward motion emanating from the center of the neuron’s receptive field. Two types of direction-selective neurons (T4/T5) are presynaptic to LPLC2 and located in the lobula plate, responsible for on and off motions respectively~\cite{bausenwein1992optic,maisak2013Drosophila}. A group of T4/T5 cells may code a specific cardinal direction and terminate in one of four neuropile of the LPLC2 to form the RMO sensibility~\cite{takemura2013visual,fischbach1989Drosophila}. Groups of LPLC2 neurons occupy the whole field of view and converge onto the giant fiber descending neurons, which trigger the jumping or take-off movement of Drosophila~\cite{card2008eacape,Reyn2017Escape,Ache2019gf}.
Fig.~\ref{fig:lplc2_model}(b) shows sketches of the anatomy of LPLC2 neurons. This indicates that the ultra-selectivity is likely to be formed by the corporation between direction-selective neurons and LPLC2, where the LPLC2 mainly integral motion information from its four preferred directions then make a judgment on radial symmetry, we assume such a process in Fig.~\ref{fig:Comparison of expansion and ROM}(b). 

However, the ultra-selectivity to radial motion opponency in Drosophila has not been widely studied in modeling works, and only a few attempts have been made to mimic the mechanism of ultra-selectivity. Zhou~\cite{zhou2022shallow} first proposes to train a Linear receptive field (LRF) based neural network with ultra-selectivity. Specifically, their model is based on Hassenstein-Reichardt correlator~\cite{hassenstein1956systemtheoretische} for acquiring directional motion information. Then, a group of LPLC2 neurons which are modeled with LRF is trained to predict the probability of collision with manually labeled looming data. Notably, their model reproduces LPLC2 neuron responses to some stimuli and discusses the two distinct solutions to inward and outward motions. Nevertheless, it is difficult to meet the real-time requirement of collision detection when the algorithm is deployed on equipment with limited computational power such as UAVs~\cite{Hector2019UAV,Loquercio2021flight}. In the same year, Hua~\cite{hua2022shaping} propose a neural network model which shapes the ultra-selectivity of LPLC2 with non-linear correlation. Similarly, the directional motion capture module also consists of multiplicative correlators. Their method can distinguish the approaching and retreating objects well, and only responds to radial motions. However, Hua's method focuses only on the detection of the expansion motion in the center of the FoV, and it was not designed for looming detection in general.

In summary, the research related to LGMD and LPLC2 has laid a foundation for our research. But capturing image motion in LGMD and the LPLC2 seems unlikely to be based on the same mechanism~\cite{gabbiani1999computation,klapoetke2017ultra,zhao2021dlgmd}, thus, we need to design a directional selective neural module for OppLoD to combine the advantages of both.

\section{Problem Formulation}
From the aspect of dynamic vision, we give the mathematical criterion of Radial Opponent Motions (ROM) in order to clarify our interested image motion. To assess the weight of importance, we propose a method to measure the degree of radial opponency between qualified image motions.

\textbf{The Radial Opponent Motions (ROM),} is a pair of image motions, which are pointing toward two opposite sides of a radius. In a real scenario, pointing along the same radius is too strict. Therefore, a small angle is tolerated, we define the following criterion for generalized judgment:
\begin{equation}
   \theta_1 - \theta_2 \in (\pi - \theta_T, \pi + \theta_T)
\end{equation}
where $\theta_1$, $\theta_2$ are the orientation of two image motions, and $\theta_T$ is the maximum angle tolerance ($\theta_T = 30^\circ$ in this paper). Vectors subject to the above criterion can always be projected to a line that is potentially to be the radius. Therefore and taken broadly, once the included angle between two vectors is within the tolerance, we consider them as a pair of ROM\footnote{Obviously, this definition allows both the inward and outward ROM case, but in this paper, only the outward case is taken into consideration since we aim to solve problems in looming detection.} and propose the measurement to describe the degree of opponency, which is called radial motion opponency (RMO).

\begin{figure}[]
  \centering
  \subfigure[]{
  \includegraphics[scale=0.33]{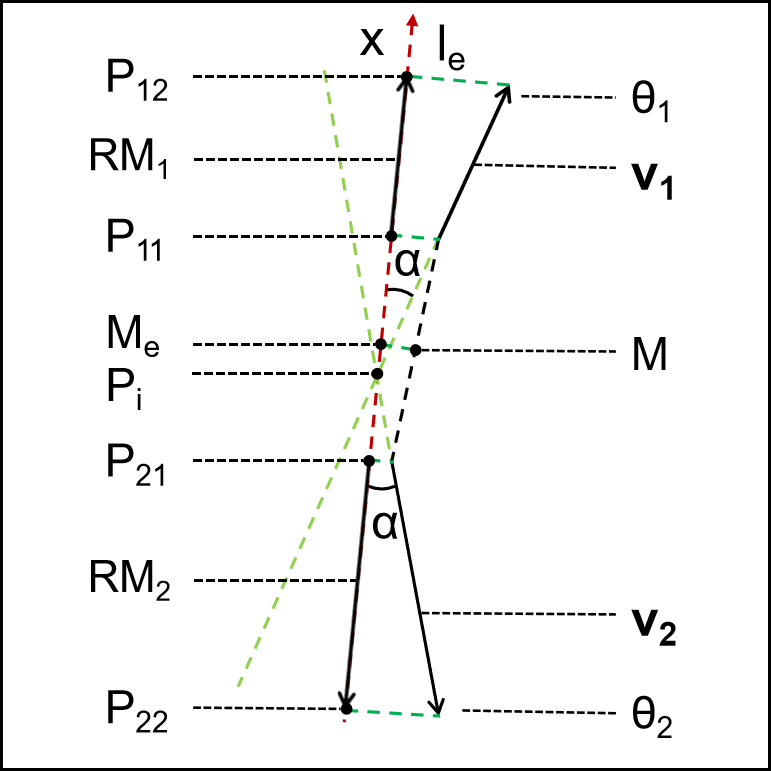}}
  \subfigure[]{
  \includegraphics[scale=0.35]{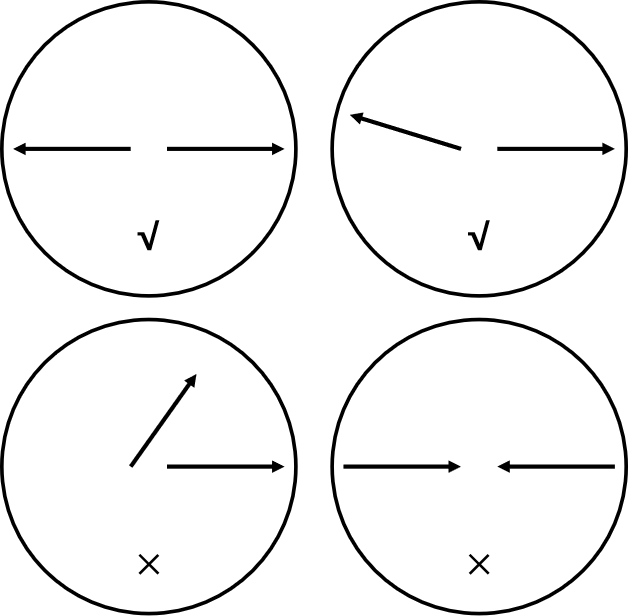}}
  \caption{The definition of RMO for a pair of conditioned vectors and the typical motion patterns. (a) The process of acquiring RMO for a pair of nonaligned vectors. First, lengthen the two vectors to intersect at a point ($P_i$). Then, draw a line ($l_e$) through $P_i$ to divide the
angle equally, and $l_e$ is the projection line (coordinate axis, x). $M_e$ represents the projection point of $M$. $P_{11}$ and $P_{21}$ are the starting points of projection, respectively. $P_{12}$ and $P_{22}$ are the ends of the projection. (b) The typical motion patterns. The cross marks represent not interested cases.}
  \label{fig:RMO}
\end{figure}

Fig.~\ref{fig:RMO} (a) describes how to project a pair of vectors, which matches the criterion, onto a radius that makes their projection symmetric from the start.
Here gives the steps:
\begin{itemize}
    \item[(1)] If there is no intersection between the vectors, extend the two vectors to make an intersection point ($P_{i}$).
    \item[(2)] Draw a line ($l_e$) through the intersection that divides the angle equally, thus, the projection angle for both vectors is the same ($\alpha$).
    \item[(3)] Project the vectors onto the ``radius", we get radial motions ($RM_1$,$RM_2$) that are strictly pointing to the opposite direction of one line.
    \item[(4)] Project the midpoint M onto the line $l_e$ and make it the center, to ensure the start points are symmetric to the center. Where M is the midpoint of the connecting line between the start points of two vectors. 
\end{itemize}

After acquiring the projected radial motion $RM_1, RM_2$, \textbf{the Radial Motion Opponency (RMO)} is defined as the ratio between the symmetric part of the projected radial motion and the sum of their magnitude:
 \begin{equation}
    \label{eq:degree of RMO}
    RMO=\frac{Length_{symmetric} (RM_{1},RM_{2})}{|RM_{1}|+|RM_{2}|}\times100\%   
\end{equation}

Where $Length_{symmetric}$ represents the length of the symmetric parts of projected motions. We propose the following method to calculate it:

Firstly, add a number axis along the radius $l_e$, make the projected midpoint $M_e$ the center.
Secendly, define two piecewise functions $f_{v1}$ and $f_{ve}$ along the number axis:
\begin{equation}
    \label{eq:function f}
    \begin{split}
    f_{v1} = \begin{cases}
                x & P_{11}\le x\le P_{12},\\
                1, & otherwise
            \end{cases}
            \\
    f_{v2} = \begin{cases}
                x & P_{21}\le x\le P_{22},\\
                1, & otherwise
            \end{cases}
    \end{split}
\end{equation}

Finally, the length of symmetric part is calculated as follows:
\begin{equation}
    \label{eq:length of symmetric part}
    Length_{symmetric}=\int_{-\infty}^{+\infty}\delta(f(x)_{v_{1}}+f(-x)_{v_{2}})dx
\end{equation}

\textbf{Thus, a ROM-sensitive neural model should response to image motions that are of high radial motion opponency (RMO).}

\section{Model Formulation}
\begin{figure*}[t]
  \centering
  \includegraphics[scale=0.6]{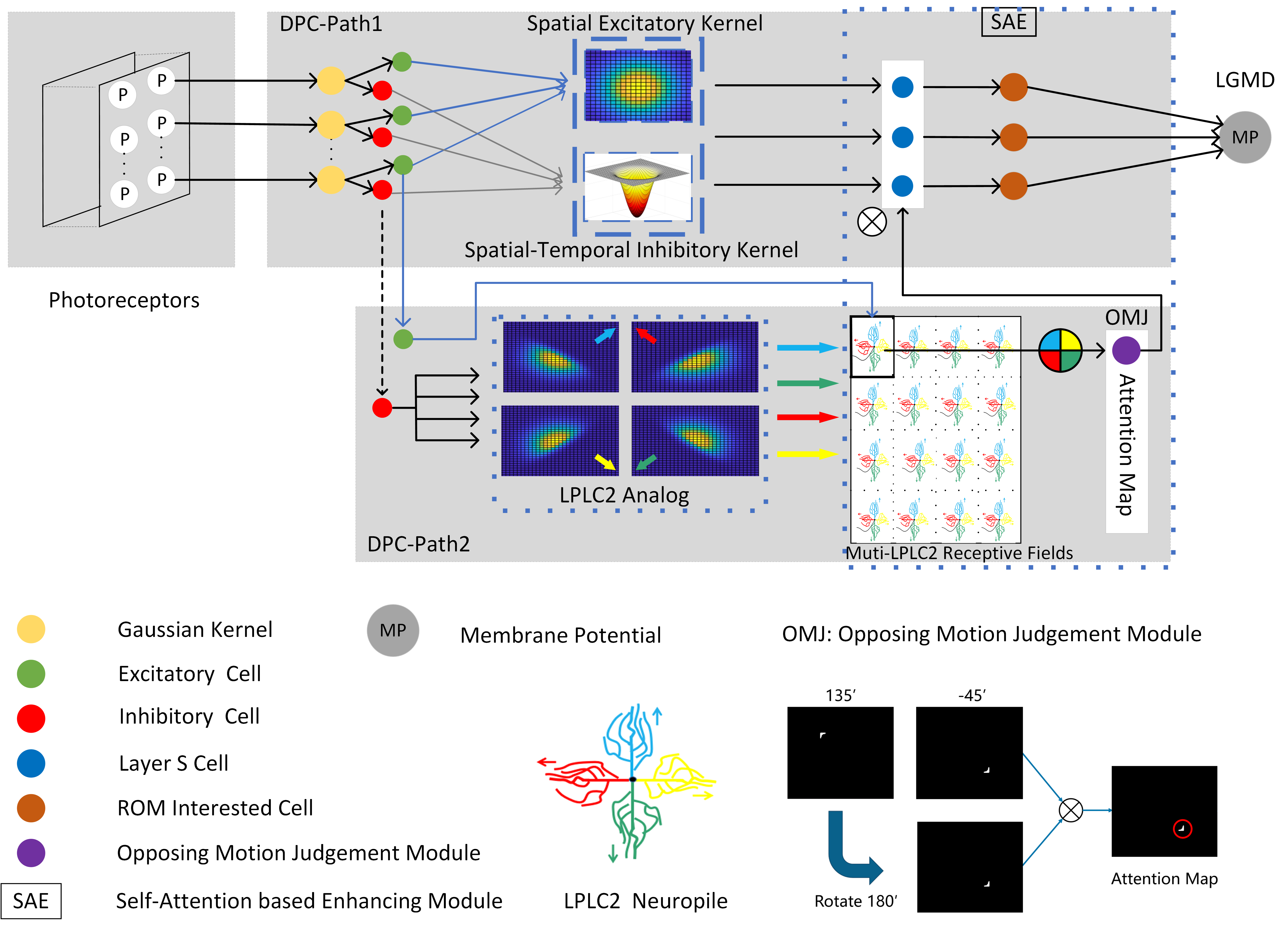}
  \caption{Schematic of our proposed neural network. There are 4 modules of our model including, the Photoreceptors, the distributed
presynaptic connection (DPC) layer, where we inserted the
motion direction extracting (MDE) module, the opposing
motion judgment (OMJ) module and the enhancing module.}
  \label{fig:Schematic}
\end{figure*}

In this section, we present the proposed neural computing model that encoded both the radial-opponent-motion (ROM) and image velocity with the inspiration of the Locust's LGMD neuron and Drosophila's LPLC2 neuropile. The schematic of the proposed model is shown in Fig.~\ref{fig:Schematic}. It consists of 4 modules, the Photoreceptors, the distributed presynaptic connection (DPC) layer, where we inserted the motion direction extracting (MDE) module, the opposing motion judgment (OMJ) module, and the Enhancing module.

\subsection{Motion Direction Extracting Module}
To endow the D-LGMD with ROM sensitivity, we need to speculate the mechanism for extracting motion direction that is compatible with D-LGMD. Luckily, although they come from different species, the main functional synaptic fans of LPLC2 and LGMD are both located in the lobula. Considering the DPC structure of the D-LGMD model is designed to extract image velocity in an ego-centric way. We designed the analog LPLC2 neuropile within the DPC structure.

The LPLC2 neurons respond intensely to outward motions emanating from the center of the neuron’s receptive field, especially when the motion is radially symmetric~\cite{maisak2013Drosophila,takemura2013visual,2015Neural,klapoetke2017ultra}. The ROM sensitivity of LPLC2 neurons is mainly based on 4 layers of neuropile where each layer encodes a preferenced motion direction respectively~\cite{klapoetke2017ultra}.

Inspired by the morphology of LPLC2, we build the motion direction extracting module with adaptive motion direction preference.

Zhao~\cite{zhao2021dlgmd} proposed the DPC structure, which defined the computational framework of synaptic interconnections in the lobula plate of the LGMD neuron. The original DPC extracts image velocity of looming targets well, but the morphology of synaptic interconnections is isotropic, thus, cannot make use of information in motion directions. In this paper, we modified the inhibitory interconnection distribution in DPC with an anisotropic kernel, to encode direction information. The motion direction extracting (MDE) module can be defined as:
\begin{equation}
    \label{eq:mde}
    \begin{split}
 D(x,y,\theta)&=\frac{1}{1+\exp( -(x'+y'))}   \\
    \begin{bmatrix}
      x' \\
      y'
    \end{bmatrix} &=
    \begin{bmatrix}
      cos\theta & -sin\theta\\
      sin\theta & cos\theta
    \end{bmatrix}
    \begin{bmatrix}
      x\\
      y
    \end{bmatrix}
    \end{split}
\end{equation}

In eq.~\eqref{eq:mde}, $\theta$ is the rotation angle. Here, $\theta$ belongs to $\left\{\left(\frac{\pi}{4}\right),\left(\frac{3\pi}{4}\right),\left(\frac{5\pi}{4}\right),\left(\frac{7\pi}{4}\right)\right\}$, corresponding to four preferred directions of LPLC2 neurons (see Fig.~\ref{fig:Schematic}). 

The novel defined inhibitory distribution defined that each pathway only suppresses a specific outward direction, which is corresponding to the four layers of LPLC2. When the four pathways converge, they cooperate to suppress inward motions toward the center. 

\subsection{Extended DPC with MDE}
In order to combine the image velocity selectivity of D-LGMD with ROM sensitivity, we propose to extending the DPC structure with MDE.  
\subsubsection{Photoreceptors Layer}
\ 

The input to DPC is luminance change, which is processed by photoreceptor (P) cells, which are formulated as:
\begin{equation}
    \label{eq:photoreceptor}
    P(x,y,t) = |L(x,y,t)-\int L(x,y,s)\delta(t-s-1)ds|
\end{equation}
In eq.~\eqref{eq:photoreceptor}, L(x,y,t) denotes the luminance value of pixel (x,y) at moment t, and $\delta$ is the unit impulse function. The photoreceptor layer prepares all the motion information without differentiating motion types.
\subsubsection{DPC Layer}
\

Below the photoreceptor layer, is the distributed presynaptic connection (DPC) layer proposed by Zhao~\cite{zhao2021dlgmd} which is a significant stage that forms the looming selectivity. This layer enhances stimuli from images of looming or high-speed moving targets and inhibits those from objects involved in lateral translation or background noise. In the DPC layer, the excitatory pathway is spatially distributed and can be described as:
\begin{equation}
    \label{eq:E layer}
    E(x,y,t)=\iint P(x,y,t)W_{E} (x-u,y-v)dudv
\end{equation}

Simultaneously, the inhibitory pathway is spatially and temporally formed and can be formulated as:
\begin{equation}
    \label{eq:P layer}
    \iiint P(x,y,t)W_{I}(x-u,y-v,t-s)dudvds
\end{equation}
In eq.~\eqref{eq:E layer} and eq.~\eqref{eq:P layer}, E(x,y,t) and I(x,y,t) are the excitation and inhibition of pixel (x,y) at moment t, respectively. $W_{E}(x,y)$, $W_{I}(x,y,t)$ represent the distribution function of excitation and inhibition. $W_{I}$ and $W_{E}$ can be in various forms to track different image velocities.

Simply, given a Gaussian kernel to describe the two distributions in the spatial domain, that is:
\begin{equation}
    \label{eq:kernel EI}
    \begin{cases}
 W_{E}(x,y)=G_{\sigma E} (x,y) \\
W_{I}(x,y,t)=G_{\sigma I}(x,y)\delta(t-\tau(x,y)) 
\end{cases}
\end{equation}
In eq.~\eqref{eq:kernel EI}, $\sigma _{E}$, $\sigma _{I}$ are the standard deviation of the
excitation and inhibition distribution. $\tau (x,y)$ denotes the temporal distribution function of the inhibitory pathway. The latency is the distance determined which increases radially with the distance extending, and can be defined as:
\begin{equation}
    \label{eq:tau}
    \tau (x,y)=\alpha+\frac{1}{\beta +\exp(-\lambda^{2}(x^{2}+y^{2}))}
\end{equation}
In eq.~\eqref{eq:tau}, $\alpha$, $\beta$ and $\lambda$ are time constants.

Next, the distributed excitation and inhibition are integrated by a linear  summation on account of they are opposing to each other, that is:
\begin{equation}
    \label{eq:original layer S}
    S(x,y,t)=E(x,y,t)-\alpha\cdot I(x,y,t)
\end{equation}
In eq.~\eqref{eq:original layer S}, $S(x,y,t)$ is the presynaptic summation of each pixel (x,y) at moment t, and $\alpha$ is the inhibition strength coefficient. In addition, because synapses stimuli are not suppressed to give negative values, a Rectified Linear Unit (ReLU) is introduced, and $S(x,y,t)$ is redefined as:
\begin{equation}
    S(x,y,t)=ReLu(S(x,y,t))
\end{equation}
where $ReLu(x)=max(0,x)$.

Following the above formulations, the DPC layer forms a temporal-spatial filter that can extract the edges from dangerous looming objects whose image on the retina moves relatively fast.

Based on the DPC manipulation, we can estimate the velocity of moving objects accurately. Furthermore, we use the motion directions extracting module introduced in section A to extract different motion directions. That is,
\begin{equation}
    \label{eq:layer OS}
    S(x,y,t,\theta)=E(x,y,t)-\alpha\cdot I(x,y,t)\cdot D(x,y,t,\theta) 
\end{equation}
In eq.~\eqref{eq:layer OS}, $\theta$ denotes motion directions.
\subsubsection{Opposing Motion Judgement Module}
\

After extracting the motion directions of moving objects, a mechanism is introduced to judge whether there are opposite motions.

For the situation that the looming object is in the center of the whole field of view (FoV), the mechanism can be formatted as:
\begin{equation}
    \label{eq:OMJ}
    S=\int_{0}^{2\pi } S_{\theta}\cdot S_{\theta+\pi} d\theta
\end{equation}
In eq.~\eqref{eq:OMJ}, $\theta$ denotes the motion direction. $S_{\theta}$ is the presynaptic sum corresponding to orientation $\theta$, $S_{\theta + \pi}$ is the opposite direction of $S_{\theta}$. Firstly, we extract the motion directions of orientation $\frac{3\pi}{4}$ and $\frac{7\pi}{4}$ respectively, obtaining $S_{\frac{3\pi}{4}}$ and $S_{\frac{7\pi}{4}}$. Next, rotate $S_{\frac{3\pi}{4}}$ 180 degrees, and multiply by $S_{\frac{7\pi}{4}}$, getting S. In the same way, rotate $S_{\frac{7\pi}{4}}$ 180 degrees, and multiply by $S_{\frac{3\pi}{4}}$. Because of the multiple operations, if there are opposing motions in the receptive field, the corresponding region will be enhanced. Similarly, we execute the judgement for $S_{\frac{\pi}{4}}$ and $S_{\frac{5\pi}{4}}$. Based on the above manipulation, the region of interest (ROI) can be acquired roughly. Furthermore, in order to remove the noise, we screen the pixels that exceed the set threshold and save the remaining part. Note that the set threshold may be various in different situations. 

More specifically, to mimic the distribution of LPLC2s, inspired by self-attention mechanism~\cite{vaswani2017attention,Goodfellow2019self-attention}, we design a novel method that can meet the situation that the looming object is not in the central of the whole FoV. First, we extract different motion directions. Then, the receptive field is zoned into some areas. Finally, judge whether there are opposite motions in each zone like the central one. In other words, we seek the correlation of different images in opposite motions locally. So, the situation that the looming object is not in the center of the whole FoV can be handled to a great extent in this way.

\subsubsection{Enhancing Module}
\

After the opposing motion judgment, the region of interest (ROI) which exists expansion can be acquired. In addition, we select the pixels aiming to remove noise. Inspired by self-attention mechanism~\cite{vaswani2017attention,Goodfellow2019self-attention}, the selected pixels are enhanced, which can be described as:
\begin{equation}
    \label{eq:layer SE}
    S_{E}(x,y,t)=S_{e}(x,y,t)\cdot S_{e}(x,y,t)\cdot c_{2}
\end{equation}
In eq.~\eqref{eq:layer SE}, $S_{e}(x,y,t)$ denotes the selected pixels that exist opposing motion, $c_{2}$ is a constant, $S_{E}(x,y,t)$ are the enhanced pixel-wise output of the enhancing module.

\section{Experiments and Results}
This section experimentally evaluated the characteristics of the proposed MDE module, and the performance of the fused neural model, from the aspects of motion-direction-selectivity, and looming-selectivity.
\subsection{Experimental setup}
Experiments are mainly conducted in recorded offline datasets, including synthetic image sequences and real videos containing the looming target.

The proposed neural network is written in MATLAB (The MathWorks, Inc., Natick, MA, USA). The computer used in the experiments is a desktop with six 3.59 GHz CPU (AMD Ryzen 5 3600) and 16 GB memory. 

\subsection{Characteristics and Effectiveness Analysis of the MDE Module}
Firstly, we evaluated the outward-motion-selectivity of the proposed MDE module with synthetic image sequences. 

\begin{figure}[]
  \centering
  \subfigure[]{
  \includegraphics[scale=0.33]{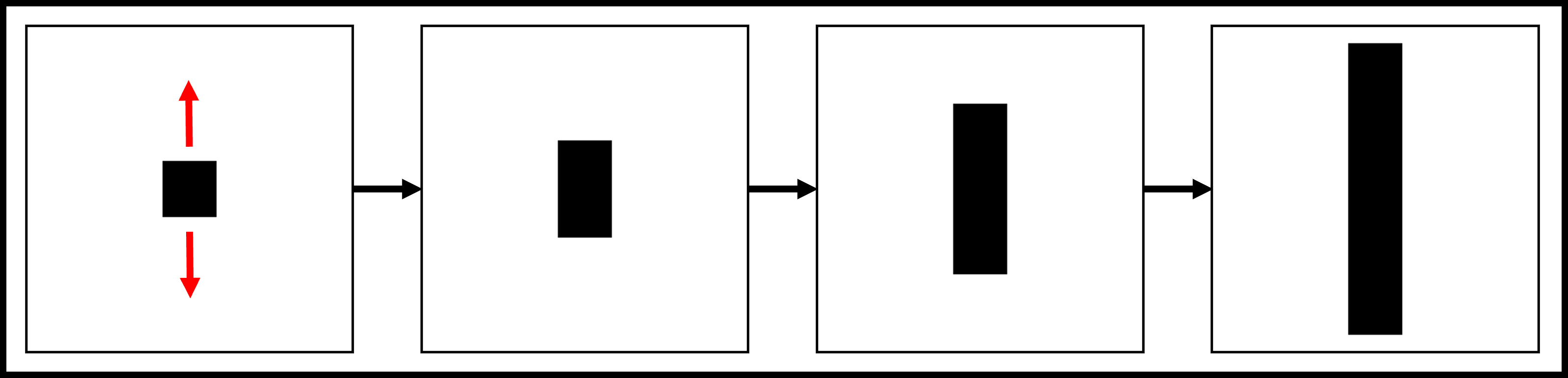}
  }
  \quad
  \subfigure[]{
  \includegraphics[scale=0.33]{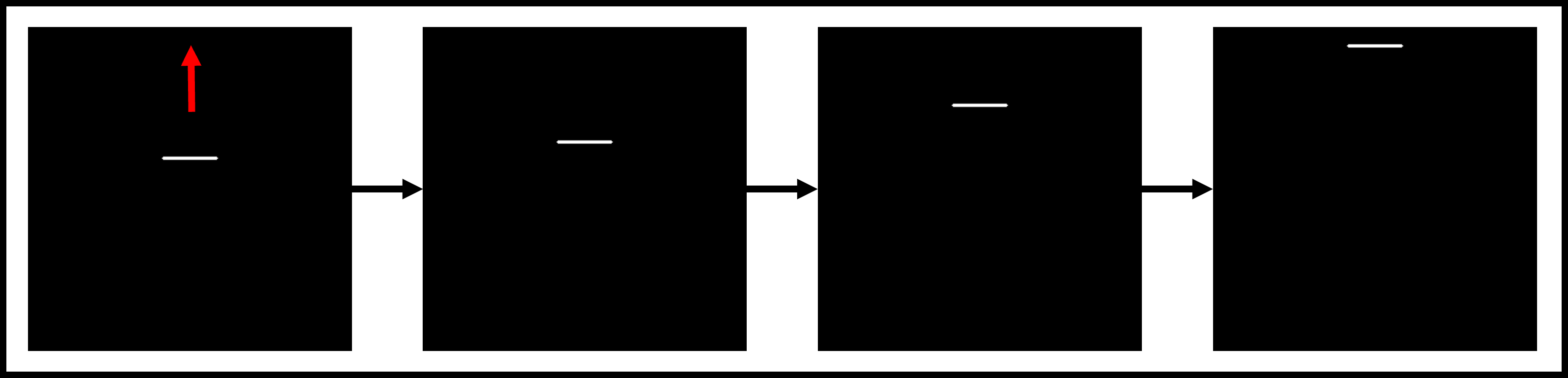}
  }
  \quad
  \subfigure[]{
  \includegraphics[scale=0.33]{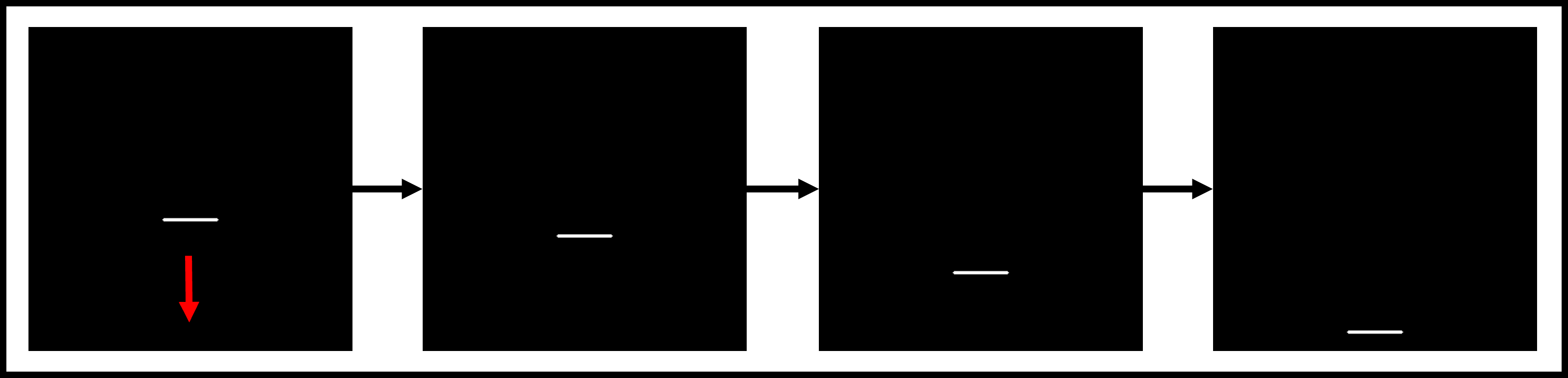}
  }
  \quad
  \subfigure[]{
  \includegraphics[scale=0.45]{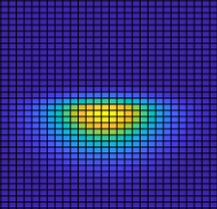}
  }
  \quad
  \subfigure[]{
  \includegraphics[scale=0.45]{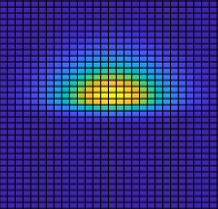}
  }
  \caption{Extracting the upward motion and downward motion. (a) A one-dimensional expanding (bar) motion towards the up and down. (b) Extracting the upward motion. (c) Extracting the downward motion. (d) and (e) Directional-selective-inhibition kernel.}
  \label{fig:MotionDirectionEvaluation}
\end{figure}
As shown in Fig.~\ref{fig:MotionDirectionEvaluation}, the input image sequences are samples of a bar expanding in the vertical direction with a constant velocity. The MDE module extracts motion towards $\frac{\pi }{2}$ and $\frac{3 \pi }{2}$ by changing the $\theta$ in eq.~\eqref{eq:mde}. Fig.~\ref{fig:MotionDirectionEvaluation}(b-c) illustrates the extracted  opposing motions, up and down, which demonstrate the MDE module can extracts a certain motion direction well. Specifically, the proposed module suppresses the opposite motion to zero with lateral inhibition, and the Directional-selective-inhibition kernel can be seen in Fig.~\ref{fig:MotionDirectionEvaluation}(d-e).

\begin{figure}[]
  \centering
  \subfigure[]{
  \includegraphics[scale=0.29]{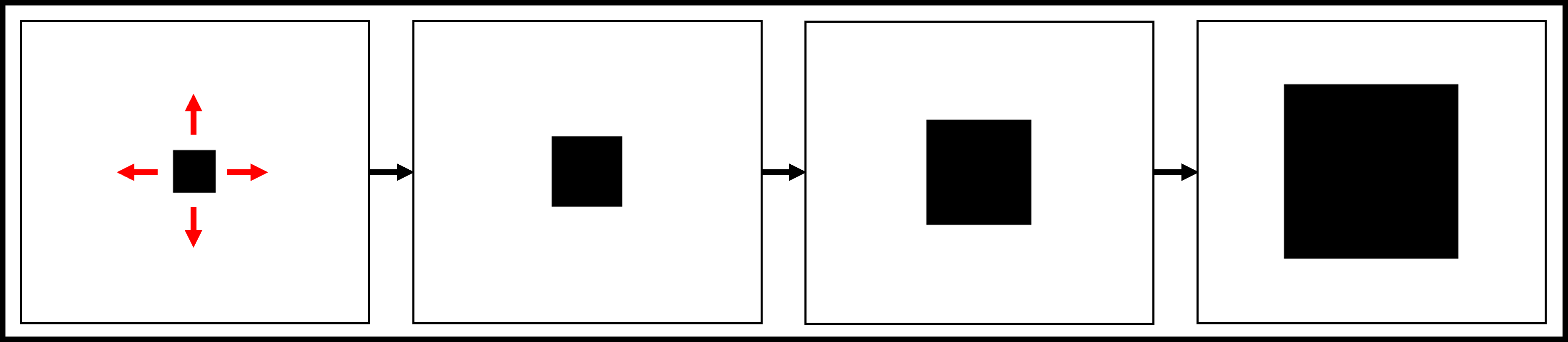}}
  \subfigure[]{
  \includegraphics[scale=0.29]{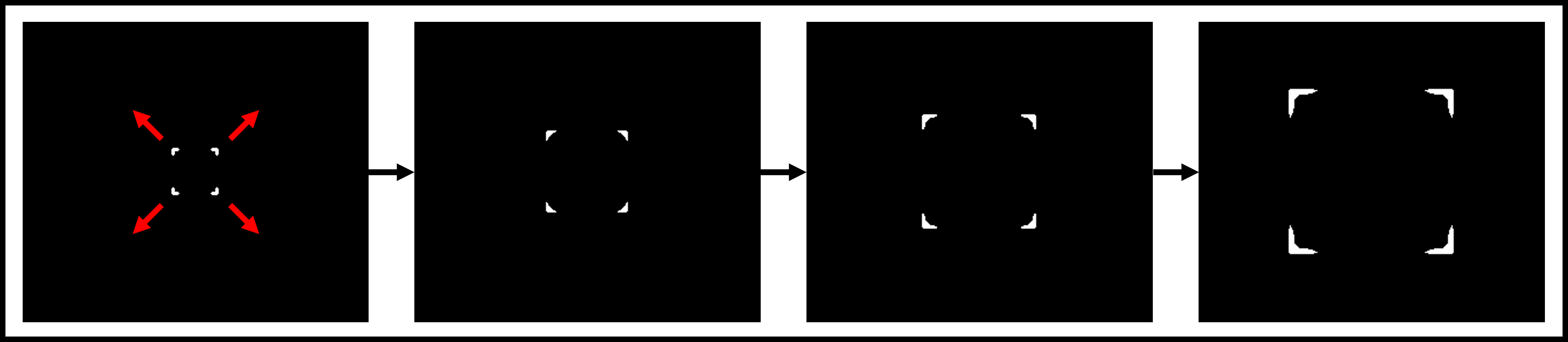}}
  \caption{Extracting the motion directions of $\frac{\pi }{4}$, $\frac{3\pi }{4}$, $\frac{5\pi }{4}$ and $\frac{7\pi }{4}$.(a) An approaching block. (b) Outputs of the MDE module. }
  \label{fig:Extracting the diagonal motions}
\end{figure}

In the LPLC2 neuron of Drosophila, outward motions in the four diagonal directions are most preferred~\cite{klapoetke2017ultra}. In the final model, we define the MDE module to follow this property. Fig.~\ref{fig:Extracting the diagonal motions} shows the diagonal motion preference, with $\theta$ set to $\frac{\pi }{4}$, $\frac{3\pi }{4}$, $\frac{5\pi }{4}$ and $\frac{7\pi }{4}$.

Figs.~\ref{fig:MotionDirectionEvaluation}(b-c) and \ref{fig:Extracting the diagonal motions}(b) demonstrate the performance of the proposed MDE module capturing the preferred motor pattern, which lays a firm foundation for further looming detection.  

\subsection{Comparison with LPLC2}
Outward motions can vary in speed, size, and symmetry. These variables have been verified by research~\cite{klapoetke2017ultra} to reveal the preference of LPLC2 neurons. In this section, in order to further study the functional properties of LPLC2 neurons, we compare the response of our model and LPLC2 neurons toward these different but similar stimuli.  

\begin{figure*}[]
  \centering
  \subfigure[]{
  \includegraphics[width=0.6\textwidth]{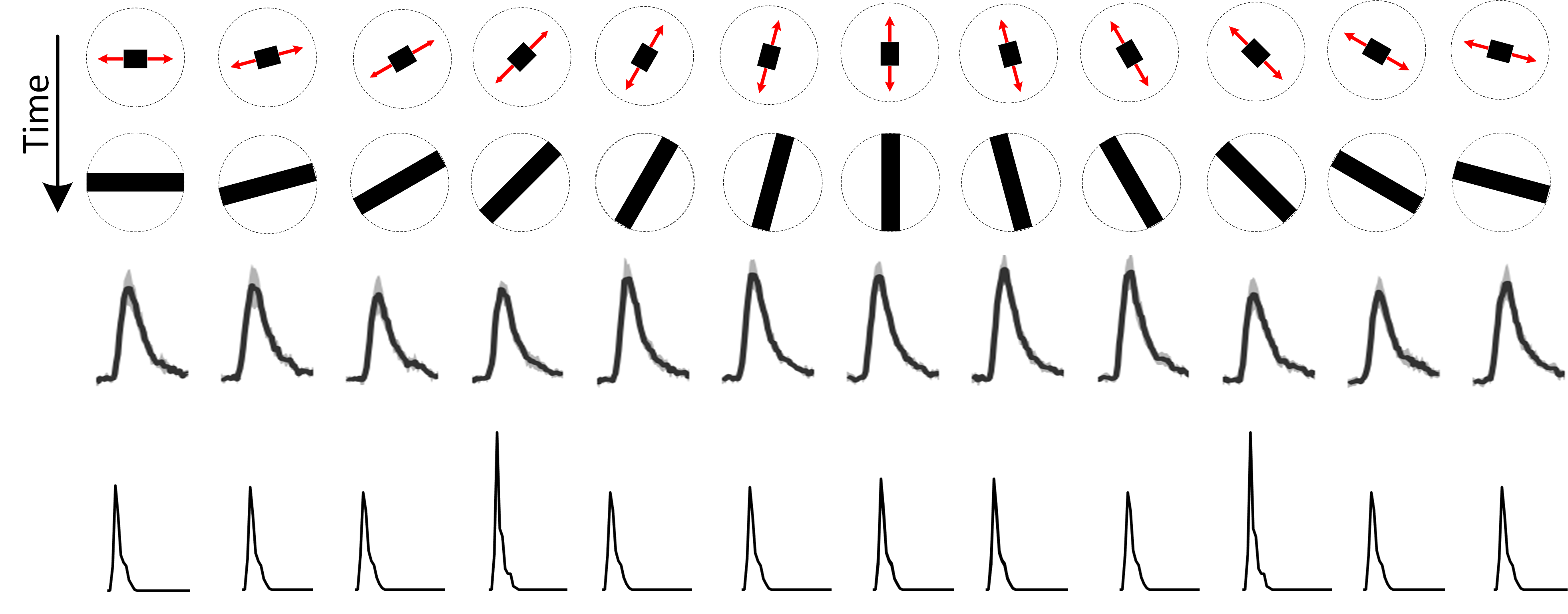}}
  \quad
  \subfigure[]{
  \includegraphics[width=0.3\textwidth]{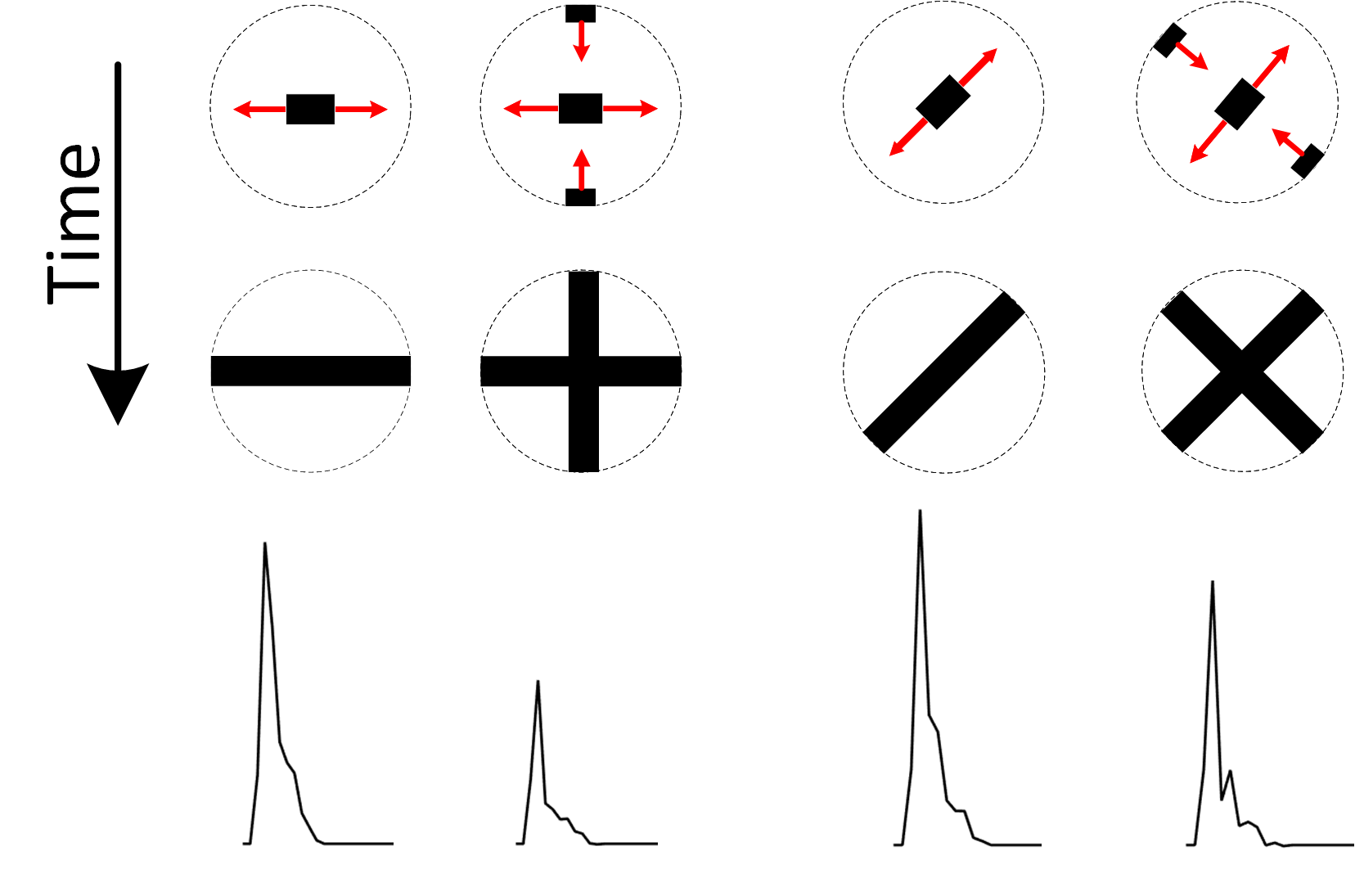}}
  \quad
  \subfigure[]{
  \includegraphics[width=0.6\textwidth]{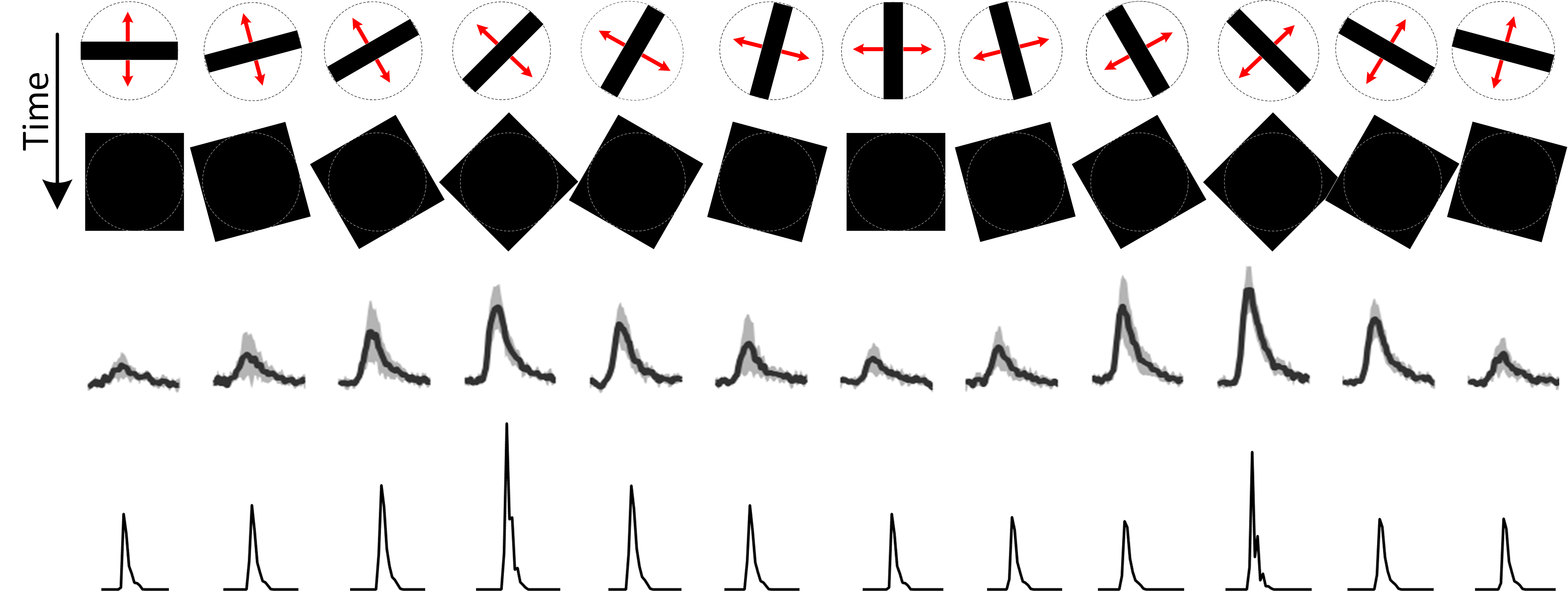}}
  \quad
  \subfigure[]{
  \includegraphics[width=0.22\textwidth]{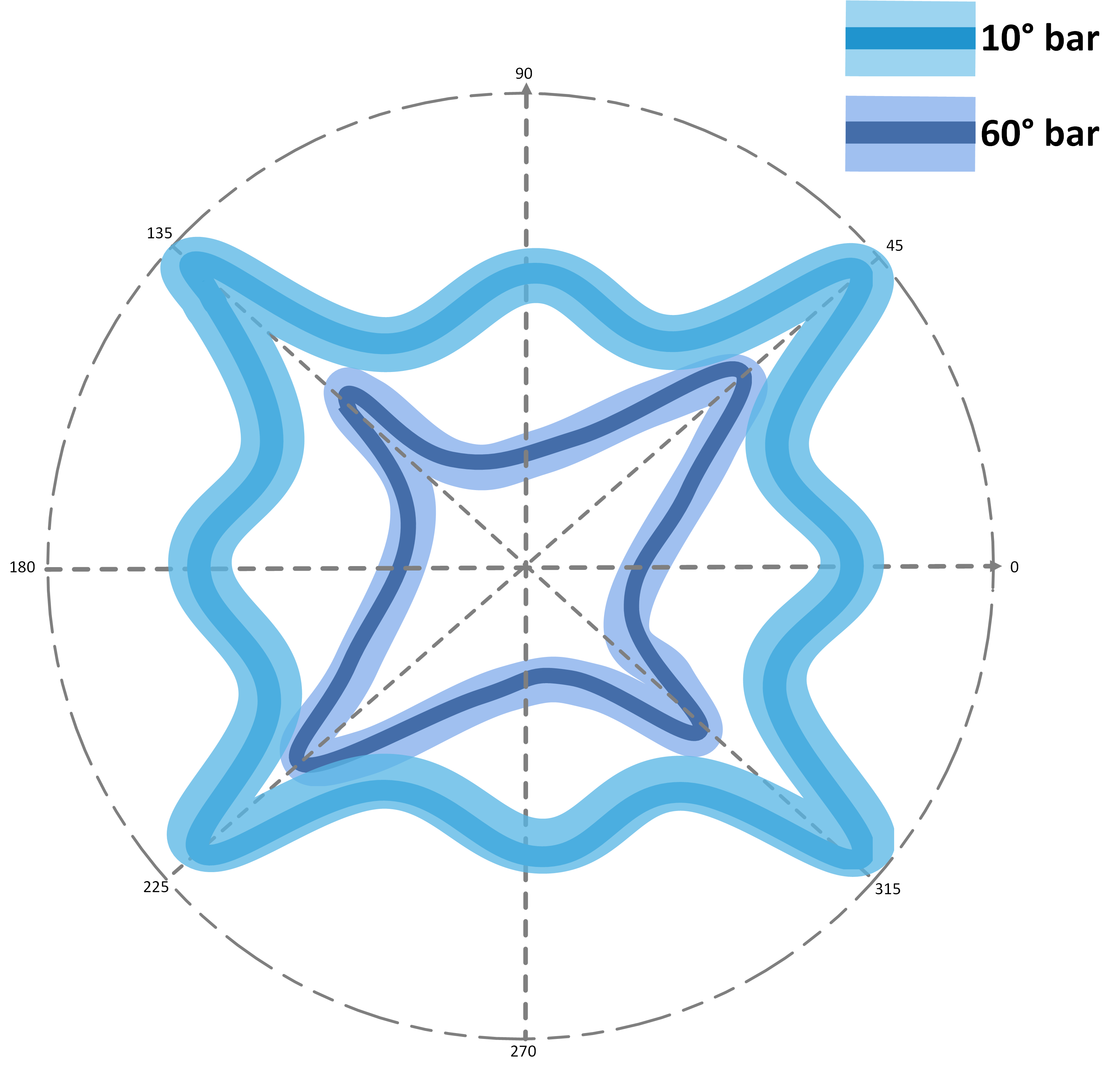}}
  \quad
  \subfigure[]{
  \includegraphics[width=0.06\textwidth]{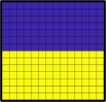}}
  \caption{Comparisons of the responses of the proposed model and LPLC2
neurons to a variety of stimuli. (a) Directional tuning with 10° one-dimensional bar expansion. (b) Responses to inward motion. (c) Directional tuning with 60° one-dimensional bar expansion. (d) Peak directional
tuning responses in polar coordinates. The relevant experimental data is from ~\cite{klapoetke2017ultra}. (e) The inhibitory filter.}
  \label{fig:Comparison with LPLC2}
\end{figure*}

Overall, the proposed model behaves quite similarly to LPLC2 neurons on many different types of stimuli, as shown in Fig.~\ref{fig:Comparison with LPLC2}. Fig.~\ref{fig:Comparison with LPLC2}(a) and (c) display the responses of expanding narrow bar(10°) and wide bar(60°), respectively. In Fig.~\ref{fig:Comparison with LPLC2}(d), the peak directional tuning responses of 10° and 60° one-dimensional expanding bar are shown in polar coordinates, which is consistent with ~\cite{klapoetke2017ultra}. In addition, We also reproduce the phenomenon of periphery inhibition that the motion signals appearing at the periphery of the receptive field suppress the activity of the LPLC2 neurons, illustrated in Fig.~\ref{fig:Comparison with LPLC2}(b).     

As we can see, the LPLC2 neurons respond to outward motions quite intensively. Moreover, there is a common characteristic of the curves in Fig.~\ref{fig:Comparison with LPLC2}(a) and (c), that is, the reactions reach maximum in the middle of the receptive field while dropping to zero at the edge. Though, there is no definite conclusion about this mechanism, at present. We infer that the T4/T5 neurons in Drosophila encode lactation information of expanding objects, which means the T4/T5 neurons only respond to the motion at specific locations (center). Based on this conjecture, our proposed model encodes the lactation information, that is, only responds to the expanding motion in the center position.
From Fig.~\ref{fig:Comparison with LPLC2}(a), we can observe that the response intensity of our model to the 10° one-dimensional
bar expansion stimuli are alike, while the response intensity of the diagonal direction is larger than others for the 10° bar, which is consistent with LPLC2 as displayed in Fig.~\ref{fig:Comparison with LPLC2}(c). Moreover, an inhibitory filter is introduced to realize periphery inhibition as illustrated in Fig.~\ref{fig:Comparison with LPLC2}(e). Following the MDE module, the inhibitory filter is applied to suppress the motion that emerges from the periphery of the receptive field and towards the center. For example, the presented filter with negative values in the upper part is to dampen the motion that appears at the upward side of the receptive field. 

\subsection{Performance for looming detection}
In this section, firstly, we use several groups of synthetic image sequences to evaluate the proposed model. Subsequently, we conduct experiments on the real datasets to further validate the performance of our model.
\subsubsection{Comparison on Synthetic Dataset}
\begin{figure}[]
  \centering
  \subfigure[]{
  \includegraphics[scale=0.286]{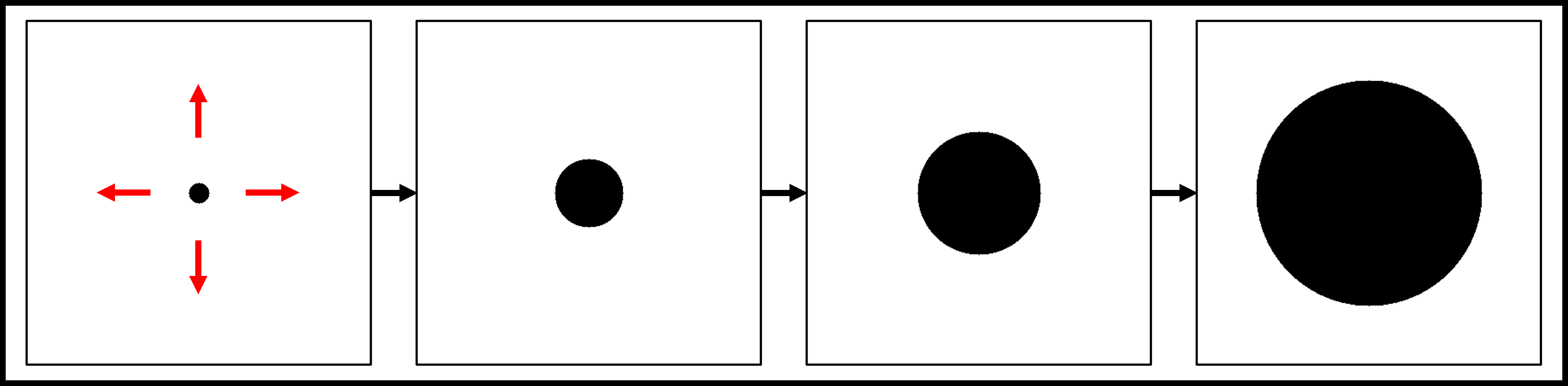}}
  \quad
  \subfigure[]{
  \includegraphics[scale=0.286]{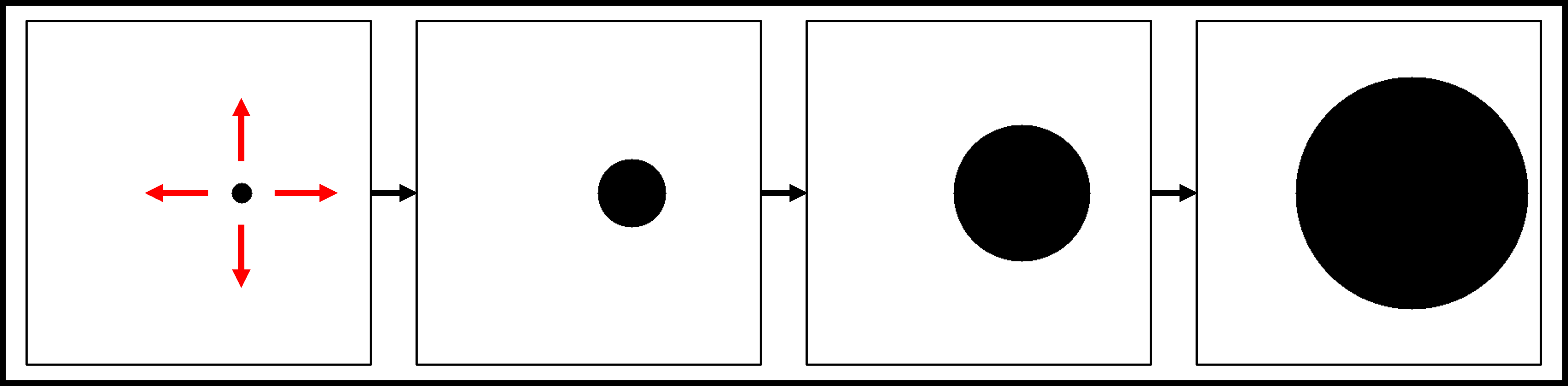}}
  \quad
  \subfigure[]{
  \includegraphics[scale=0.208]{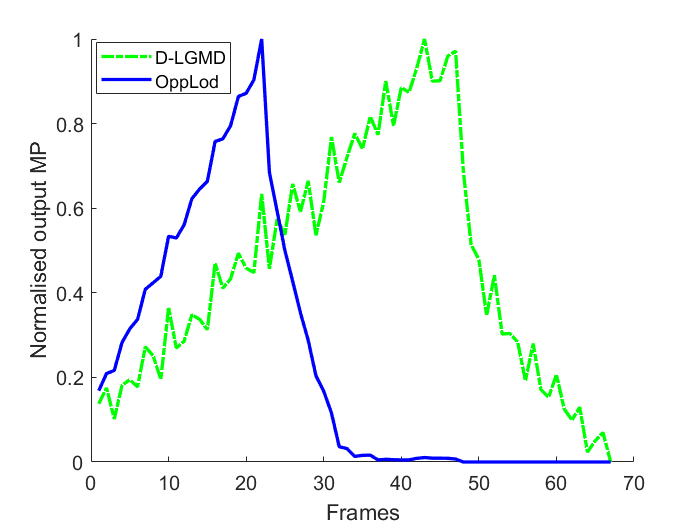}}
  \quad
  \subfigure[]{
  \includegraphics[scale=0.208]{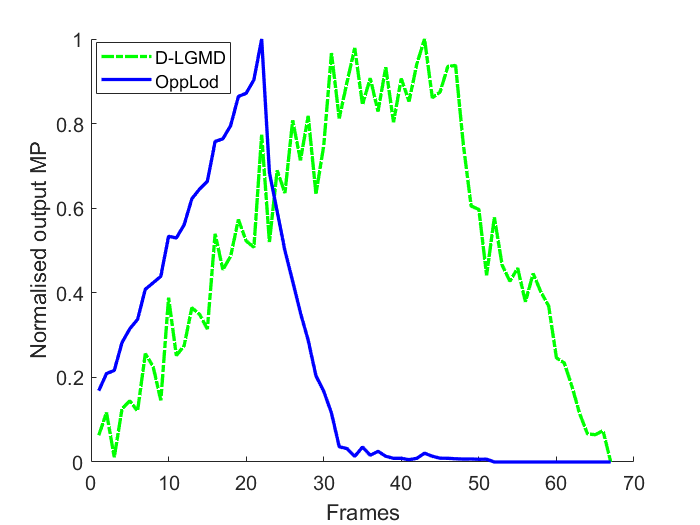}}
  \quad
  \subfigure[]{
  \includegraphics[scale=0.26]{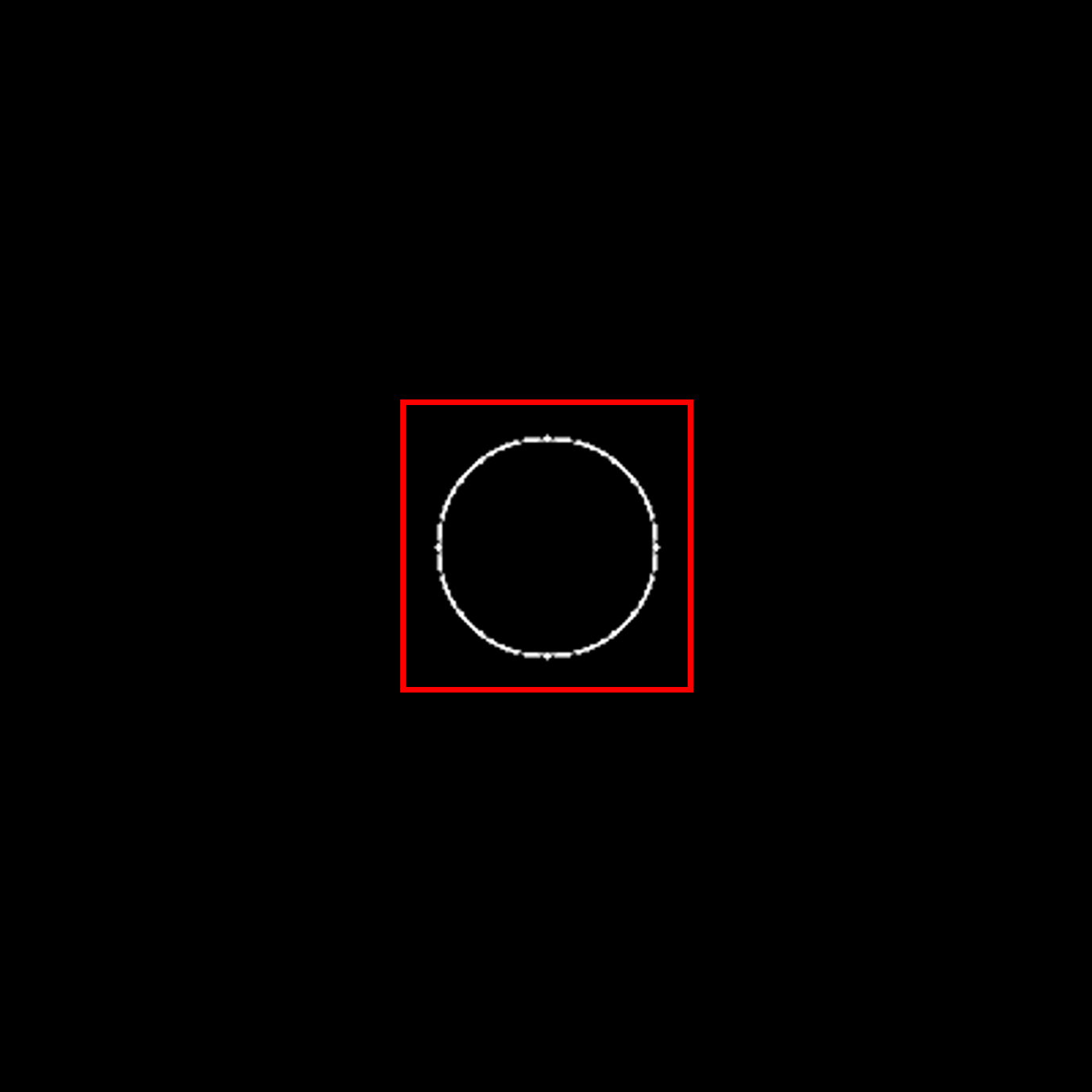}}
  \quad
  \subfigure[]{
  \includegraphics[scale=0.26]{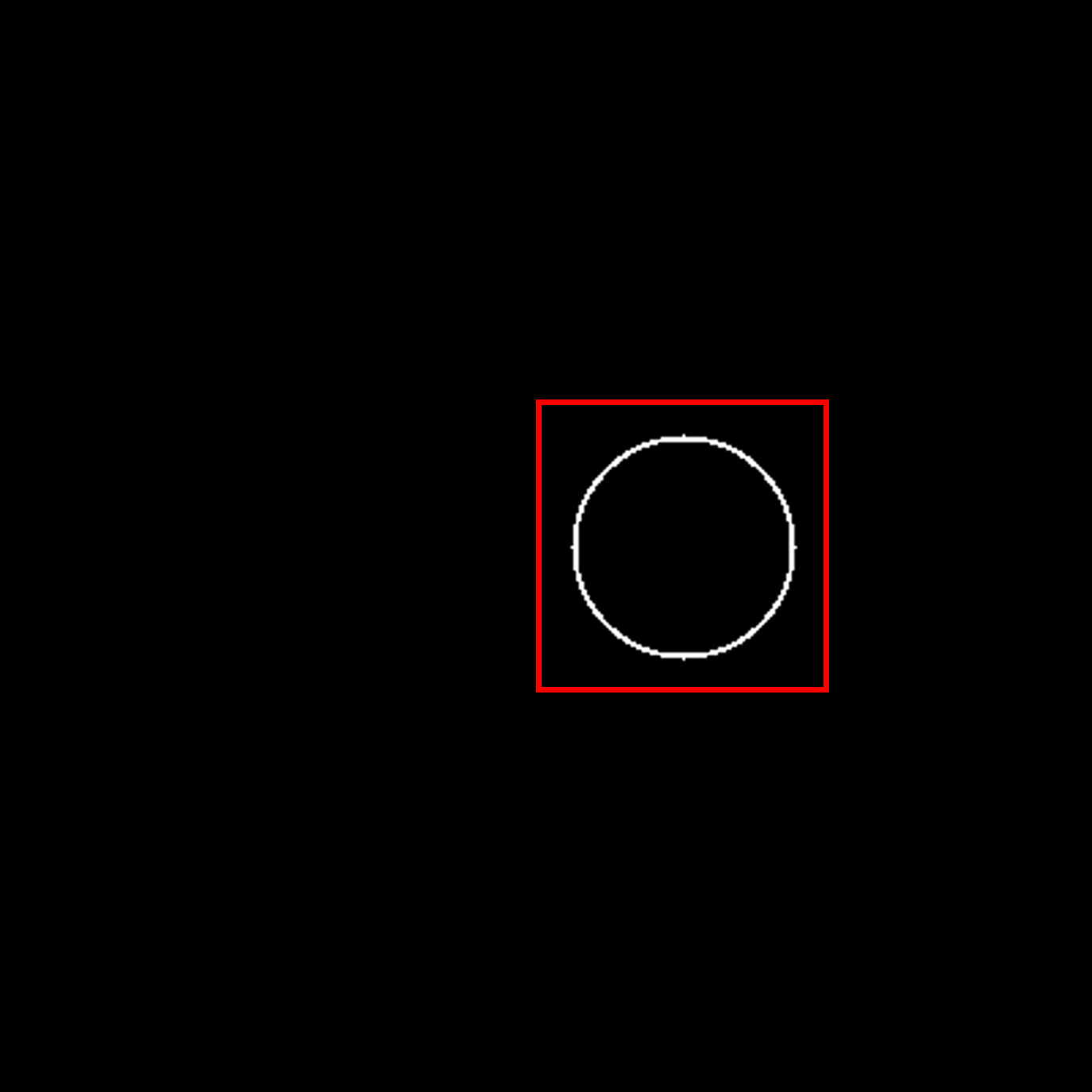}}
  \caption{Responses of the proposed model and D-LGMD to looming disk. (a)-(b) An approaching disk located in the center and non-center. (c)-(d) Outputs of the proposed model and D-LGMD. (e)-(f) Sampled the  object.}
  \label{fig:Looming disk}
\end{figure}

The input scene is a looming black disk that can be categorized into two types, center and non-center. The motion of the center represents that the looming object is in the center of the whole FoV, and the motion of the non-center is the situation the looming object is not in the center of the whole FoV. The input synthetic image sequences of looming events are displayed in Fig.~\ref{fig:Looming disk}(a)-(b).

As shown in Fig.~\ref{fig:Looming disk}(c)-(d), we compare the normalized output between D-LGMD and our model in the same scene. Generally, our proposed model gets better performance than the baseline model in both two situations. In Fig.~\ref{fig:Looming disk}(c), for the motion generated from the center of the whole FoV, our model's output reaches the peak at near frame 22, while the baseline model at near frame 43. As displayed in Fig.~\ref{fig:Looming disk}(d), for the motion beginning off-center, our model also performs much better. In Fig.~\ref{fig:Looming disk}(e), the detected object is shown in the red box.

The output curves also illustrate that both two models respond to the looming motion at the beginning, but our proposed model has a higher sensitivity than the baseline, due to the introduction of the ROM feature. It can be seen that our model responds to the looming objects stronger and earlier, which means that can attain more time avoiding obstacles such as for UAV. Note that the performance for looming detection of the proposed model is relevant to the number and size of LPLC2 neurons. Here, the number of the model unit is 25, and the size of the receptive field is set to 200 × 200. Additionally, our model can not only detect the approaching obstacle but also localize it, which has great potential applications in object detection, drone landing, and so on.

\subsubsection{Comparison on Real Dataset}
\begin{figure}[]
  \centering
  \subfigure[]{
  \includegraphics[scale=0.35]{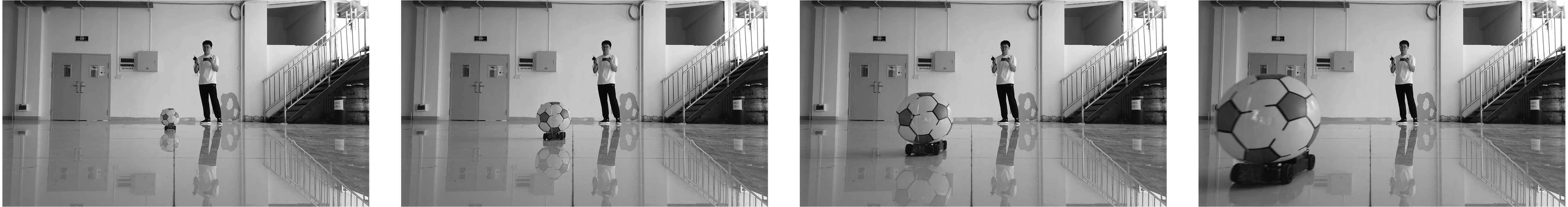}}
  \quad
  \subfigure[]{
  \includegraphics[scale=0.35]{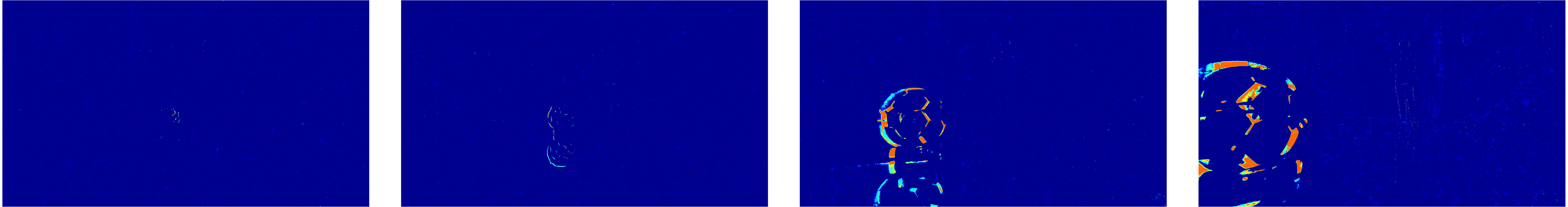}}
  \quad
  \subfigure[]{
  \includegraphics[width=0.45\textwidth]{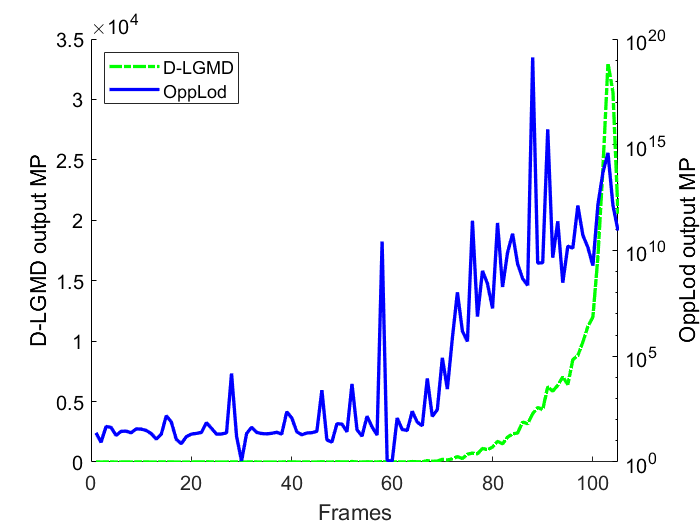}}
  \caption{Comparative responses of OppLod and D-LGMD neural networks with the object approaching the camera in the real scene. (a) Input image sequence. (b) The output of the proposed model. (c) Response curves of D-LGMD and our model. Note that the ball is installed on a remote control toy car.}
  \label{fig:ball_enhancement}
\end{figure}

\
We further test our proposed model with real-world video sequences in an indoor environment where the objects, e.g., balls, cubes, and the camera.      
Fig.~\ref{fig:ball_enhancement} presents the performance of a single ball approaching the camera. As displayed in Fig.~\ref{fig:ball_enhancement}(a), the ball starts moving from the center of the FoV, and approaches the edge of the image. The detected object is shown in Fig.~\ref{fig:ball_enhancement}(b). From the response curves of the two models, we can see that the proposed model yields a better detection performance
than the baseline model. As illustrated in Fig.~\ref{fig:ball_enhancement}(c), the proposed model begins to sense the moving ball near frame 63, and its curve reaches the maximum at frame 88. For the baseline model, start noticing the object near frame 88, and reaches the peak at frame 103. 
Note that, because of the ball's own shaking as it accelerates at frame 58, the proposed model has a relatively large output, shown in Fig.~\ref{fig:ball_enhancement}(c). Though the proposed model works better than the baseline model in sensing the looming object, there still remains some noise. Hence, a further accurate estimation may be needed for avoiding risk in practical applications, such as actual flight.  

\begin{figure}[]
  \centering
  \subfigure[]{
  \includegraphics[scale=0.35]{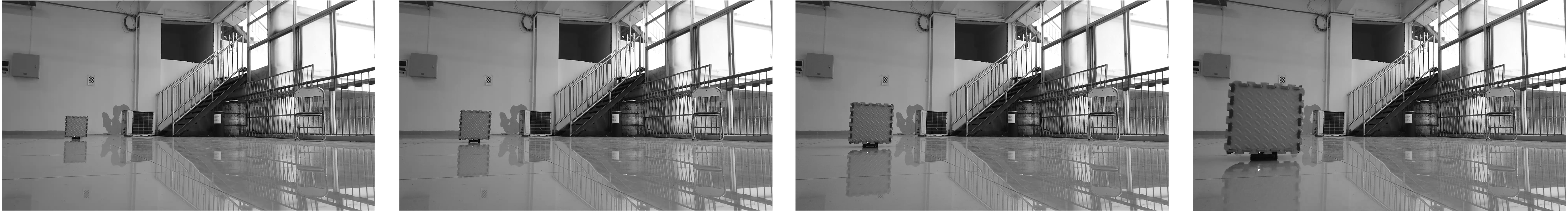}}
  \quad
  \subfigure[]{
  \includegraphics[scale=0.35]{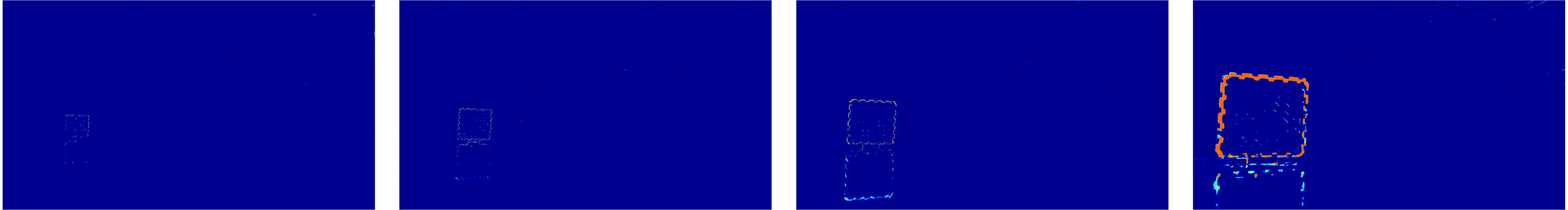}}
  \quad
  \subfigure[]{
  \includegraphics[scale=0.43]{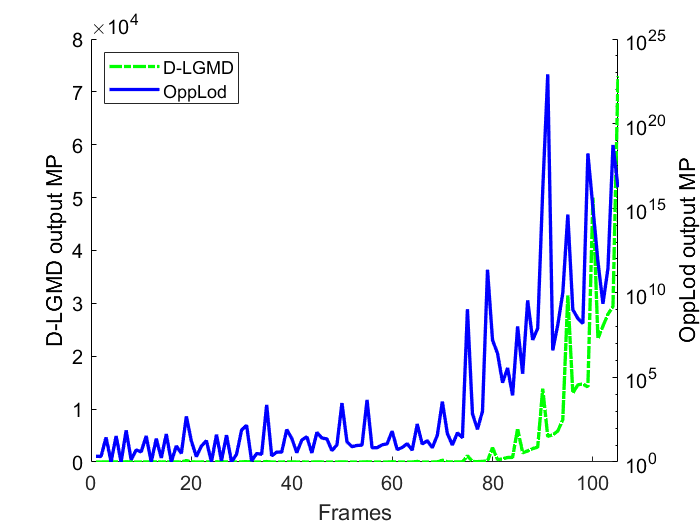}}
  \caption{Responses compared with D-LGMD for the looming object lying on one side of the field of view. (a) Input image sequence. (b) The output of the proposed model. (c) Response curves of D-LGMD and our model. Note that the cube is also installed on a remote control toy car.}
  \label{fig:cube_noncenter}
\end{figure}

As shown in Fig.~\ref{fig:cube_noncenter}, we evaluate the proposed model with a cube that lies on the left side of the FoV approaching the camera. From Fig.~\ref{fig:cube_noncenter}(b), it can be seen that the proposed model has a good performance for the scene in that the looming obstacle lies on one side of the FoV, near the edge. Additionally, it works well even the looming object is comparatively small. In Fig.~\ref{fig:cube_noncenter}(c), the response curve of the proposed model reaches the top at frame 91 while the baseline model is at frame 105. It is clear that the proposed model reacts to the danger earlier. Moreover, due to the structure that the LPLC2 neurons distributed over the field of view, our proposed model can easily detect the approaching object near the edge of the FoV.

Meanwhile, from  Fig.~\ref{fig:ball_enhancement}(b) and Fig.~\ref{fig:cube_noncenter}(b), we can see that the proposed model also highlights the shadow of the approaching object which is influenced by light. In future work, visual neurons which eliminate the shadow will be integrated together to enhance the detection performance.

\begin{figure}[]
  \centering
  \subfigure[]{
  \includegraphics[scale=0.35]{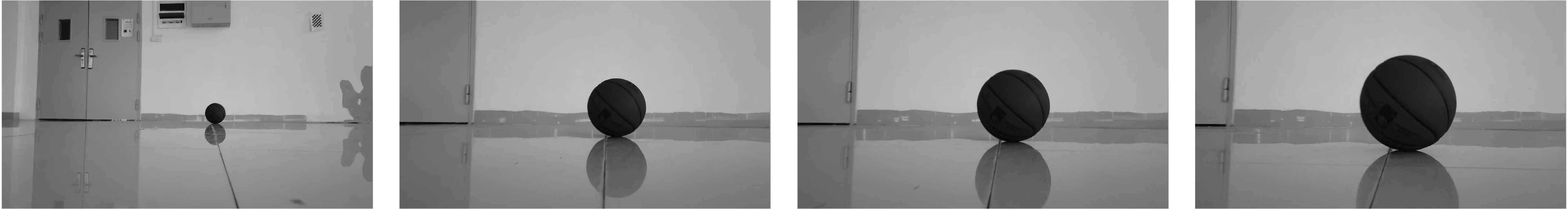}}
  \quad
  \subfigure[]{
  \includegraphics[scale=0.35]{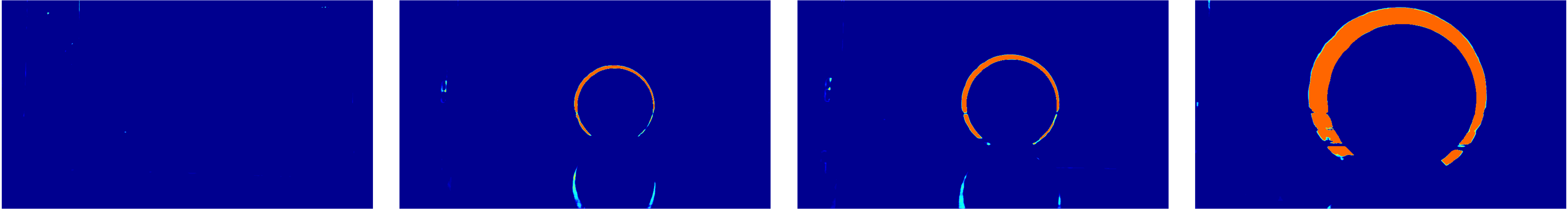}}
  \quad
  \subfigure[]{
  \includegraphics[width=0.45\textwidth]{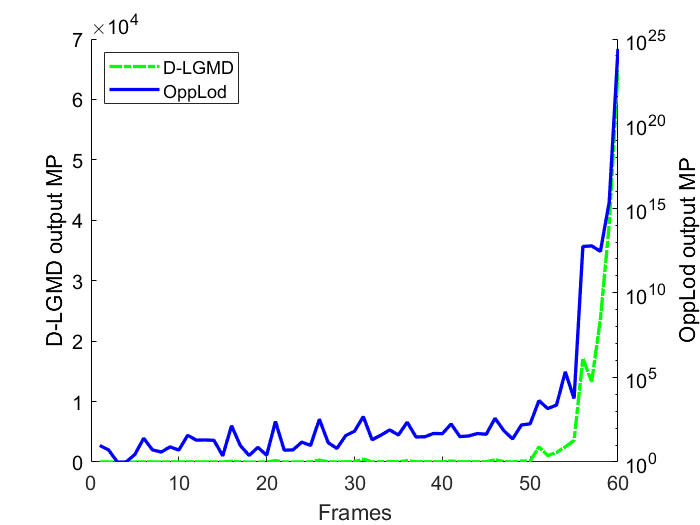}}
  \caption{Responses of the proposed model and D-LGMD neural networks with the camera approaching the object in the real scene. (a) Input image sequence. (b) The output of the proposed model. (c) Response curves of the proposed model and D-LGMD.}
  \label{fig:ball_cameramove}
\end{figure}

Fig.~\ref{fig:ball_cameramove} displays the movements of the camera, involving the approach of basketball. A similar scene can be seen in Fig.~\ref{fig:ball_enhancement}, but the moving one is the camera. On the whole, the proposed model can accomplish the detecting task well and respond to the risk timely. From the neural response curves of the two models, we can observe that the proposed model outperforms the baseline model for sensing the looming detection.     

\section{Conclusions}

In this paper, we have proposed a visual model fused with LGMD and LPLC2 for looming detection. To better describe the pattern of our interested image motion, the Radial Opponent Motion (ROM), we provide a mathematical criterion and a method to measure its degree. Our proposed model involves the ROM-sensitive structure based on lateral inhibition, which mimics the function of the four layers of synaptic fans in the LPLC2. The effectiveness of the lateral-inhibition-based motion direction extracting (MDE) module is verified with synthetic image sequences. The lateral-inhibition-based motion direction extracting (MDE) module exhibited clear motion direction preference and thus provide materials for radial motion detection. The proposed model shows consistency with the LPLC2 neurons for a variety of stimuli related to RMO. When applied to real scenario videos, the proposed model has a good perception of approaching risk, with advantages in robustness and sensibility. In future work, we will further explore the potential application of the proposed ROM-sensitive model for high-speed collision detection such as in UAV scenarios.

\appendix




\bibliographystyle{model1-num-names}

\bibliography{cas-refs}



\bio{Figures/Feng_Shuang}
\textbf{Feng Shuang} dean and professor of the School of Electrical Engineering, Guangxi University. He once worked at the Institute of Intelligent Machines (IIM), and the University of Science and Technology of China (USTC), as the director of the Robot Sensor Laboratory. He got a Bachelor's degree from Special Class of Gifted Young, USTC, in 1995, and a Ph.D. degree from 2004 to 2009. In 2009, he joined IIM as a full professor and then was selected as a member of “One Hundred Talented People of Chinese Academy of Sciences”. His research focuses on intelligent robots, multi-dimensional force sensors, quantum system control, etc. He has published more than 60 papers and applied for more than 10 national invention patents.
\endbio

\bio{Figures/Yanpeng_Zhu}
\textbf{Yanpeng Zhu} received the B.S. degree in Automation from the School of Electrical and Electronic Engineering, Shijiazhuang Tiedao University, Shijiazhuang, China, in 2019. Now he is currently pursuing the M.S. degree from the School of Electrical Engineering, Guangxi University, Nanning, China. His current research interests include bio-inspired visual algorithms and deep learning.
\endbio

\vspace{3\baselineskip}

\bio{Figures/Yupeng_Xie}
\textbf{Yupeng Xie} received the B.S degree in Electronics Science and Technology from the School of Physics, Guangxi University, Nanning, China, in 2022. Now he is currently pursuing the M.S. degree from the School of Electrical Engineering, Guangxi University, Nanning, China. His current research interests include bio-inspired visual algorithms and autonomous landing of drones.
\endbio

\vspace{3\baselineskip}

\bio{Figures/Lei_Zhao}
\textbf{Lei Zhao} received his bachelor's degree in measurement and control technology and instruments from Jiangsu University of Technology in 2017. At present, he is studying for a master's degree in electrical engineering at the School of Electrical Engineering, Guangxi University, Nanning, China. His current research interests include event cameras and visual algorithms.
\endbio

\bio{Figures/Quansheng_Xie}
\textbf{Quansheng Xie} received the B.S. degree in electrical engineering and automation from the School of Electrical and Information, Northeast Agricultural University, Harbin, China, in 2021. Now he is currently pursuing the M.S. degree from the School of Electrical Engineering, Guangxi University, Nanning, China. His current research interests include bio-inspired visual algorithms and dynamic vision sensor.
\endbio

\vspace{3\baselineskip}

\bio{Figures/JiannanZhao}
\textbf{Jiannan Zhao} received the B.Eng. degree from the College of Opt-Electronics Engineering, Chongqing University, Chongqing, China, in 2016, and the Ph.D. degree in computer science from the University of Lincoln, Lincoln, U.K., in 2021. He was a Research Assistant with Tsinghua University, Beijing, China, from 2017 to 2018. He is currently an Assistant Professor with the School of Electrical Engineering, Guangxi University, Nanning, China. His current research interests include autonomous UAVs, bio-inspired visual algorithms, and artiﬁcial intelligence.
\endbio

\vspace{3\baselineskip}

\bio{Figures/Shigang_Yue}
\textbf{Shigang Yue} received the B.Eng. degree from the Qingdao University of Technology, Qingdao, China, in 1988, and the M.Sc. and Ph.D. degrees from the Beijing University of Technology (BJUT), Beijing, China, in 1993 and 1996, respectively. He was with BJUT as a Lecturer from 1996 to 1998 and an Associate Professor from 1998 to 1999. He was an Alexander von Humboldt Research Fellow with the University of Kaiserslautern, Kaiserslautern, Germany, from 2000 to 2001. He is currently a Professor of computer science with the School of Computer Science, University of Lincoln, Lincoln, U.K. Before joining the University of Lincoln as a Senior Lecturer in 2007 and promoted to a Reader in 2010 and a Professor in 2012, he held research positions with the University of Cambridge, Cambridge, U.K., Newcastle University, Newcastle upon Tyne, U.K., and University College London, London, U.K., respectively. His current research interests include artiﬁcial intelligence, computer vision, robotics, brains and neuroscience, biological visual neural systems, evolution of neuronal subsystems, and their applications, e.g., in collision detection for vehicles, interactive systems, and robotics. Dr. Yue is a member of the International Neural Network Society, International Society of Artiﬁcial Life, and International Symposium on Biomedical Engineering. He is the Founding Director of the Computational Intelligence Laboratory (CIL), Lincoln, and the Deputy Director of Lincoln Center for Autonomous Systems (L-CAS), Lincoln. He is also a Co-Founding Director of the Machine Life and Intelligence Research Center, Guangzhou, China. He is the Coordinator for several EU FP7/Horizon2020 Projects.
\endbio

\end{document}